\documentclass[10pt,twocolumn,letterpaper]{article}
\PassOptionsToPackage{table}{xcolor}
\usepackage[pagenumbers]{iccv} 
\usepackage{bm}
\newcommand{\mname}{Amodal3R\xspace}
\definecolor{iccvblue}{rgb}{0.21,0.49,0.74}
\usepackage[pagebackref,breaklinks,colorlinks,allcolors=iccvblue]{hyperref}

\makeatletter
\renewcommand{\paragraph}{%
  \@startsection{paragraph}{4}%
  {\z@}{-0.0em}{-0.5em}%
  {\normalfont\normalsize\bfseries}%
}
\makeatother

\renewcommand{\footnotemark}{\dagger}

\def\paperID{2040}
\def\confName{ICCV}
\def\confYear{2025}

\title{\mname: Amodal 3D Reconstruction from Occluded 2D Images}

\author{
    Tianhao Wu$^{1*}$, Chuanxia Zheng$^{2\footnotemark}$, Frank Guan$^3$, Andrea Vedaldi$^2$, Tat-Jen Cham$^1$\\
    \vspace{-8pt}\\
    $^*$S-Lab, $^1$Nanyang Technological University,~
    $^2$Visual Geometry Group, University of Oxford\\
    \vspace{-12pt}\\
    $^3$Singapore Institute of Technology\\
    \vspace{-12pt}\\
    {\tt\small \{tianhao001,astjcham\}@ntu.edu.sg, \{cxzheng,vedaldi\}@robots.ox.ac.uk}\\
    \vspace{-16pt}\\
    {\tt\small Frank.guan@singaporetech.edu.sg}
}

\begin{document}

\twocolumn[{
\maketitle
\vspace*{-1.0cm}
\begin{center}
\includegraphics[width=\textwidth]{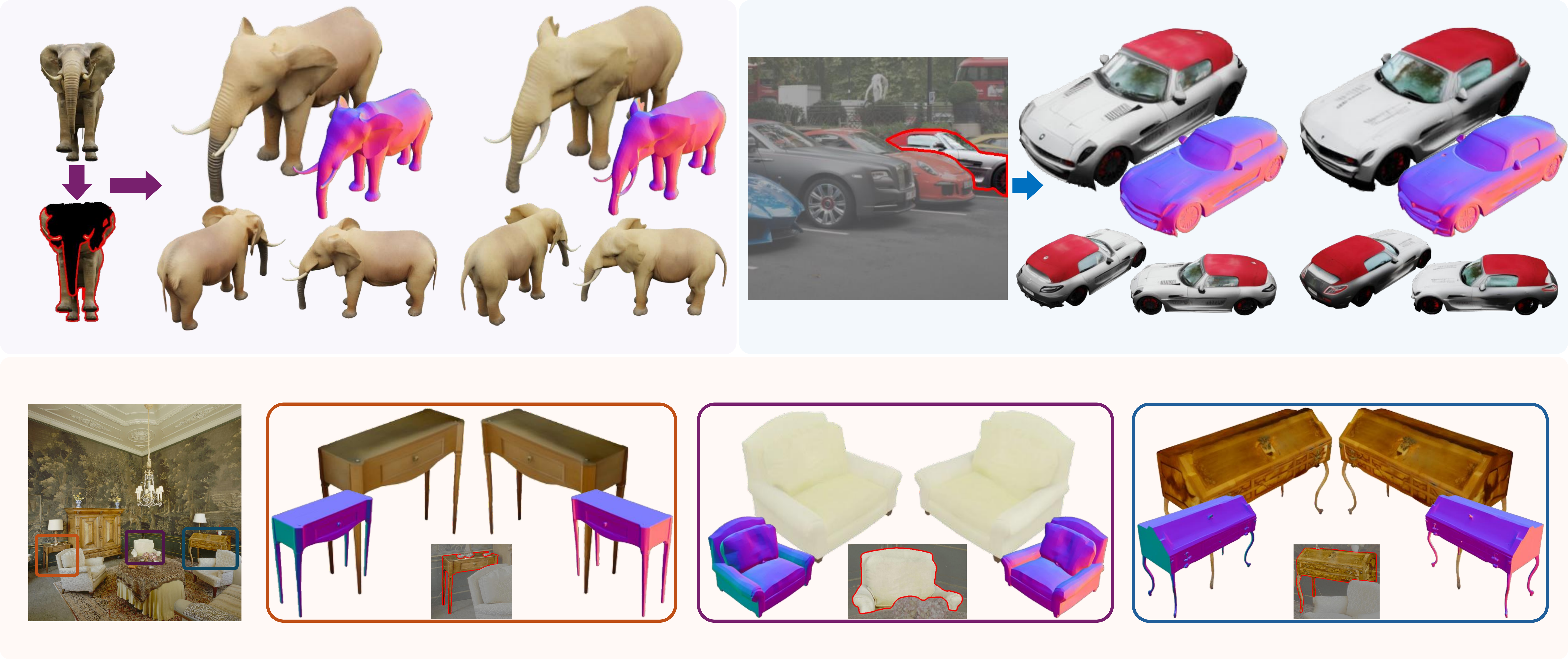}
\begin{picture}(0,0)  
    \put(-244,210){\normalsize \textbf{3D Recast}}
    \put(-240,111){\footnotesize \textbf{2D Input}}
    \put(-185,111){\footnotesize \textbf{Stochastic sample 1}}
    \put(-87,111){\footnotesize \textbf{Stochastic sample 2}}
    \put(-10,210){\normalsize \textbf{In-the-wild Reconstruction}}
    \put(16,111){\footnotesize \textbf{2D Input}}
    \put(88,111){\footnotesize \textbf{Stochastic sample 1}}
    \put(177,111){\footnotesize \textbf{Stochastic sample 2}}
    \put(-244,97){\normalsize \textbf{3D Scene Decomposition}}
    \put(-224,16.5){\footnotesize \textbf{2D Input}}
    \put(-2,16.5){\footnotesize \textbf{Decomposed 3D assets}}
\end{picture}
\vspace{-12pt}
\captionsetup{hypcap=false}
\captionof{figure}{\textbf{Example results of \mname.}
Given partially visible objects within images (occluded regions are shown in black, visible areas in red outlines),
our \mname generates \emph{diverse} semantically meaningful 3D assets with reasonable geometry and plausible appearance. \emph{We sample multiple times to get different results from the same occluded input.}
Trained on synthetic datasets, it generalizes well to real-scene and in-the-wild images, where most objects are partially visible, and reconstructs reasonable 3D assets.
}
\label{fig:teaser}
\end{center}
}]
\footnotetext[2]{Project Lead.}
\begin{abstract}
Most image-based 3D object reconstructors assume that objects are fully visible, ignoring occlusions that commonly occur in real-world scenarios.
In this paper, we introduce \mname, a conditional 3D generative model designed to reconstruct 3D objects from partial observations.
We start from a ``foundation'' 3D generative model and extend it to recover plausible 3D geometry and appearance from occluded objects.
We introduce a mask-weighted multi-head cross-attention mechanism followed by an occlusion-aware attention layer that explicitly leverages occlusion priors to guide the reconstruction process.
We demonstrate that, by training solely on synthetic data, \mname learns to recover full 3D objects even in the presence of occlusions in real scenes.
It substantially outperforms existing methods that independently perform 2D amodal completion followed by 3D reconstruction, thereby establishing a new benchmark for occlusion-aware 3D reconstruction. 
See our project page \href{https://sm0kywu.github.io/Amodal3R/}{\textcolor{red}{https://sm0kywu.github.io/Amodal3R/}}.
\end{abstract}
    
\section{Introduction}%
\label{sec:intro}

Humans possess a remarkable ability to infer the complete 3D shape and appearance of objects from single views, even when the objects are partly hidden behind occluders.
This ability,
namely \emph{amodal completion},
is key to operating in complex real-world environments, where objects are often partially occluded by surrounding clutter.
However, existing image-based 3D reconstruction models~\cite{nguyen2019hologan,nash2020polygen,chan2022efficient,watsonnovel,Szymanowicz_2023_ICCV,liu2023zero,zheng2024free3d,voleti2025sv3d,hong2023lrm,chen2024lara,siddiqui2024meshgpt,szymanowicz2024splatter,zhang2024clay,xiang2024structured} fail to recover full 3D assets when the object is partially occluded.
We thus consider the problem of reconstructing 3D objects from one or more partially-occluded views, a novel task that we call \emph{amodal 3D reconstruction}.

Amodal 3D reconstruction is a challenging task that
requires reconstructing an object's 3D geometry and appearance while completing its occluded parts, both of which are highly ambiguous.
Previous approaches to amodal 3D reconstruction~\cite{ozguroglu2024pix2gestalt,chen2024partgen} have decomposed the task into 2D amodal completion~\cite{zhan2020self,zheng2021visiting,zhan2024amodal}, followed by conventional 3D reconstruction~\cite{liu2023zero,zheng2024free3d,voleti2025sv3d}.
While these two-stage pipelines are easy to implement, they have some drawbacks.
First, 2D amodal completion methods rely predominantly on appearance-based priors rather than explicit 3D structural cues.
This \emph{lack of 3D geometric reasoning} is suboptimal.
Second, some 3D reconstruction methods can use or require multiple views to function.
In this case, 2D amodal completion may lack \emph{multi-view consistency}, particularly when it is performed independently for different views, which confuses the 3D reconstruction process.

In this paper, we introduce \textbf{\mname}, a novel \emph{single-stage} paradigm for amodal 3D reconstruction that surpasses previous state-of-the-art approaches (see \cref{fig:teaser}).
\mname augments the ``foundation'' 3D generator TRELLIS~\cite{xiang2024structured} with an additional branch to condition on occlusions.
Its key advantage is performing reconstruction, completion, and occlusion reasoning directly in a 3D latent space instead of using a \emph{two-stage} scheme.
In this way, the model can reconstruct both visible and occluded regions of the object coherently and plausibly.
To adapt TRELLIS to amodal reconstruction, we introduce \emph{mask weighting} in multi-head cross-attention and a new \emph{occlusion-aware} layer.
These guide the model to focus more on visible parts of the object without perturbing the pre-trained model too much.

We evaluated the effectiveness of \mname on datasets like Google Scanned Objects~\cite{downs2022google} and Toys4K~\cite{stojanov2021using} augmented with occlusions, on 3D scenes from Replica~\cite{straub2019replica},
and on real-world monocular images.
Without relying on additional heuristics, \mname achieves significantly superior performance compared to state-of-the-art models and generalizes well to different datasets, including real ones.

In summary, our main contributions are as follows:
\begin{itemize}
    \item We propose a novel occlusion-aware 3D reconstruction model that directly reconstructs complete and high-quality 3D objects from one or more partially occluded views, without relying on 2D amodal completion models.
    \item We introduce a mask-weighted cross-attention mechanism and an occlusion-aware layer to inject occlusion awareness into an existing 3D reconstruction model, improving both the geometry and appearance of the reconstructed objects when they are partially occluded.
    \item We demonstrate via experiments on the 3D scene dataset Replica and real-world images that our one-stage pipeline significantly outperforms existing two-stage ones, establishing a new benchmark for amodal 3D reconstruction.
\end{itemize}
\section{Background}%
\label{sec:background}

\paragraph{2D Amodal Completion.}

Recent advances in 2D amodal completion~\cite{zhan2020self,zheng2021visiting,ao2024open,ozguroglu2024pix2gestalt,zhan2024amodal,xu2024amodal} have achieved significant success in reconstructing occluded regions of objects in 2D images.
While these methods show promise for 3D generation pipelines~\cite{ao2024open,ozguroglu2024pix2gestalt}, they still have limitations.
First, 2D amodal completion models have limited 3D geometric understanding as they interpret images as 2D patterns.
While excelling at completing textures, they may generate physically implausible geometries when handling complex occlusions.
This stems from their lack of explicit 3D reasoning and reliance on 2D appearance priors, without true volumetric understanding.
Second, for models that use multi-view images as input, the results of the 2D amodal completion are often inconsistent across views.
Inconsistent views need to be corrected by the 3D reconstructor, which cause confusion (see \cref{sec:exp}).
Although there has been significant work on multi-view consistent generation~\cite{Szymanowicz_2023_ICCV,zheng2024mvd, xie2024carve3d, shi2023mvdream, tang2024mvdiffusion++,chen2024partgen}, multi-view consistent completion is less explored.
RenderDiffusion~\cite{anciukevivcius2023renderdiffusion} contains an example, but the results are often blurry or lack details.
This motivates our 3D-centric reconstruction framework.

\begin{figure*}[tb!]
  \centering
  \includegraphics[width=\linewidth]{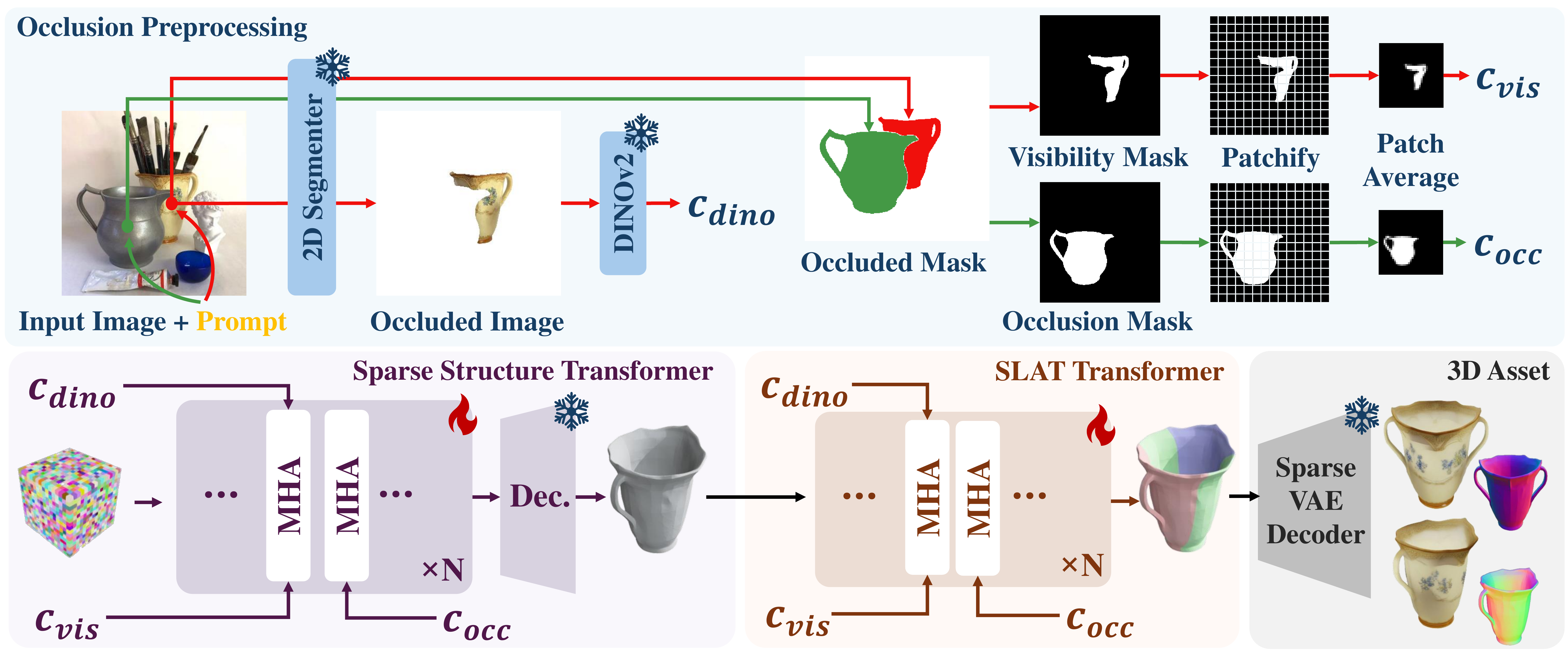}
  \vspace{-14pt}
  \caption{\textbf{Overview of \mname.}
  Given an image as input and the \textcolor[RGB]{255,192,0}{regions of interest}, \mname first extracts the partially visible target object, along with the \textcolor[RGB]{240,16,12}{visibility} and \textcolor[RGB]{68,153,69}{occlusion} masks using an off-the-shelf 2D segmenter.
  It then applies DINOv2~\cite{oquabdinov2} to extract features $\bm{c}_{dino}$ as additional conditioning for the 3D reconstructor.
  To enhance occlusion reasoning, each transformer block incorporates a \emph{mask-weighted cross-attention} (via $\bm{c}_{vis}$) and \emph{occlusion-aware attention layer} (via $\bm{c}_{occ}$),
  ensuring the 3D reconstructor accurately perceives visible information while effectively inferring occluded parts.
  For conditioning details, see \cref{sec:mask-weighted-cross-attn}.
  }
  \label{fig:overview}
\end{figure*}

\paragraph{3D Shape Completion.}

Several methods~\cite{stutz2018learning, zhou20213d,chu2024diffcomplete, cui2024neusdfusion} start from a partial 3D reconstruction, then complete it in 3D.
They use encoder-decoder architectures~\cite{hu20203d} or volumetric representations~\cite{cheng2023sdfusion,kasten2024point} to robustly recover 3D shape but not texture, which is necessary in many applications.
They also still require recovering the partial 3D geometry from an occluded image, a challenge in itself.
Furthermore, they ignore the input image when completing the object in 3D, disregarding the occlusion pattern that caused the 3D geometry to be recovered only partially.
Recent work~\cite{cho2024robust} utilizes the partially visible object as input specifically for 3D shape completion. However, it focuses solely on geometry reconstruction, without recovering the object's texture and appearance.
In contrast, our approach is end-to-end, recovering the complete 3D shape and appearance of the object from the occluded image.

\paragraph{3D Generative Models.}

Early advancements in 3D generation are based on GANs~\cite{goodfellow2020generative}, exploring various 3D representations such as point clouds~\cite{lin2018learning,huang2020pf}, voxel grids~\cite{yang20173d,zhu2018visual}, view sets~\cite{park2017transformation,nguyen2019hologan}, NeRF~\cite{schwarz2020graf,chan2021pi,niemeyer2021giraffe,chan2022efficient}, SDF~\cite{gao2022get3d}, and 3D Gaussian mixtures~\cite{wewer2024latentsplat}.
While GANs can capture complex 3D structures, they struggle with stability, scalability, and data diversity.
The focus then shifted to diffusion models~\cite{sohl2015deep,ho2020denoising,rombach2022high}, which were more capable and versatile.
They were first applied to novel view synthesis~\cite{watson2022novel} in image space, before expanding to model a variety of 3D representations, including
point clouds~\cite{luo2021diffusion,nichol2022point,wu2023sketch,melas2023pc2,zheng2024point},
voxel grids~\cite{li2023diffusion,muller2023diffrf}, triplanes~\cite{shue20233d,zou2024triplane}, meshes~\cite{gupta20233dgen,long2024wonder3d}, and 3D Gaussian mixtures~\cite{li2023gaussiandiffusion,lan2023gaussian3diff,zhang2024gaussiancube,chen2024mvsplat360}.
Autoregressive models~\cite{nash2020polygen,siddiqui2024meshgpt,chen2024meshanything} have also been explored for mesh generation, focusing on improving the topology of the final 3D mesh.

Authors have also shifted from performing diffusion in 2D image space~\cite{shi2023mvdream,Szymanowicz_2023_ICCV,zheng2024free3d} to 3D latent spaces~\cite{vahdat2022lion,jun2023shap,ntavelis2023autodecoding,xiang2024structured,roessle2024l3dg,shuang2025direct3d}.
Such methods typically consist of two stages: the first for generating geometry and the second for generating appearance, and are capable of high-quality image-to-3D generation.
However, they assume that input objects are fully visible, which limits their application to real-world scenes, where occlusions are frequent.
Here, we extend such approaches to generate high-quality 3D assets from occluded input images --- a challenging task that requires inferring complete 3D objects from partial information.
\section{Method}%
\label{sec:method}

Consider an image $x$ containing a partially occluded view of an object of interest $\bm{o}_i$,
we wish to reconstruct the \emph{complete} 3D shape and appearance $y$ of the object $\bm{o}_i$.
This task is thus conceptually similar to existing image-to-3D reconstruction, except that here the object is partially occluded instead of being fully visible in $x$.
We call this problem \emph{amodal 3D reconstruction}.

Here,
we elaborate on \mname (\cref{fig:overview}), a new method for amodal 3D reconstruction.
Unlike recent two-stage methods~\cite{ozguroglu2024pix2gestalt,chen2024partgen} that first perform 2D amodal completion followed by 3D reconstruction,
\mname is an end-to-end occlusion-aware 3D reconstruction model that generates \emph{complete} shapes and multi-view geometry directly within the 3D space.
Formally, \mname is a conditional generator $\upsilon_\theta(y|o_i,M_\text{vis},M_\text{occ})$ that takes as input the image $x$, centered on the partially visible object $\bm{o}_i$, as well as the visibility mask $M_\text{vis}$ and occlusion mask $M_\text{occ}$.
The visibility mask $M_\text{vis}$ marks the pixels of image $x$ containing the object $\bm{o}_i$,
while $M_\text{occ}$ marks the pixels containing the occluders, \ie, all other objects that potentially obscure $\bm{o}_i$.
For real images,
these masks can be efficiently obtained using pre-trained 2D segmentation models like Segment Anything~\cite{kirillov2023segment}.
By providing point coordinate prompts for the interest object
$\bm{o}_i$
and its occluders respectively,
the segmentation model can generate the corresponding masks.
In cases where multiple occluders are present or when occluders fragment the visible area of the target object, a sequential process is employed. Specifically, point prompts for each visible/occluding region are provided to the 2D segmenter one at a time, with the model generating an individual mask for each region.
These masks are then aggregated to form a comprehensive visibility/occlusion mask.

The challenges for \mname are how to:
\textbf{(1)} produce a \emph{plausible} and \emph{complete} 3D shape and appearance from partial observations, even in the absence of multi-view information; and
\textbf{(2)} ensure 3D \emph{consistency} in terms of geometry and photometry, maintaining seamless visual coherence between visible and generated regions.

\subsection{Preliminaries: TRELLIS}%
\label{sec:preliminaries}

We begin by briefly reviewing the TRELLIS~\cite{xiang2024structured} model on which our model is based.
TRELLIS is a conditional 3D diffusion model that performs denoising in a sparse 3D latent space.
First, it introduces a transformer-based variational autoencoder (VAE) ($\mathcal{E},\mathcal{D}$), where the encoder $\mathcal{E}$ maps sparse voxel features to structured latents $\bm{z}$, and the decoder $\mathcal{D}$ converts them into desired output representations, including 3D Gaussians~\cite{kerbl20233d}, radiance fields, and meshes.
In particular, a 3D object $\bm{o}_i$ is encoded using its \emph{structured latent variables} (SLAT) defined as 
$
\bm{z} = \{(\bm{z}_i, \bm{p}_i)\}_{i=1}^{L}
$,
where $\bm{z}_i \in \mathbb{R}^C$ is a local latent feature attached to the voxel at position $\bm{p}_i \in \{0, 1, \dots, N - 1\}^3$, $N$ is the spatial resolution of the grid, and $L \ll N^3$ is the number of active voxels intersecting the object's surface.
This representation encodes both coarse geometric structures and fine appearance details by associating local latents with active voxels.

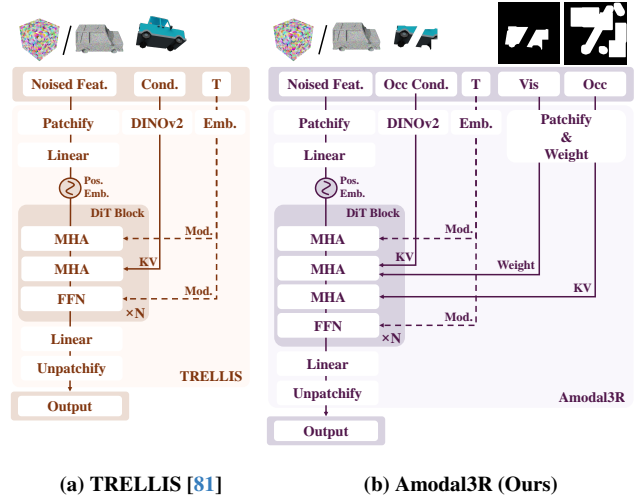
\begin{figure}[tb!]
    \centering
    \includegraphics[width=\linewidth]{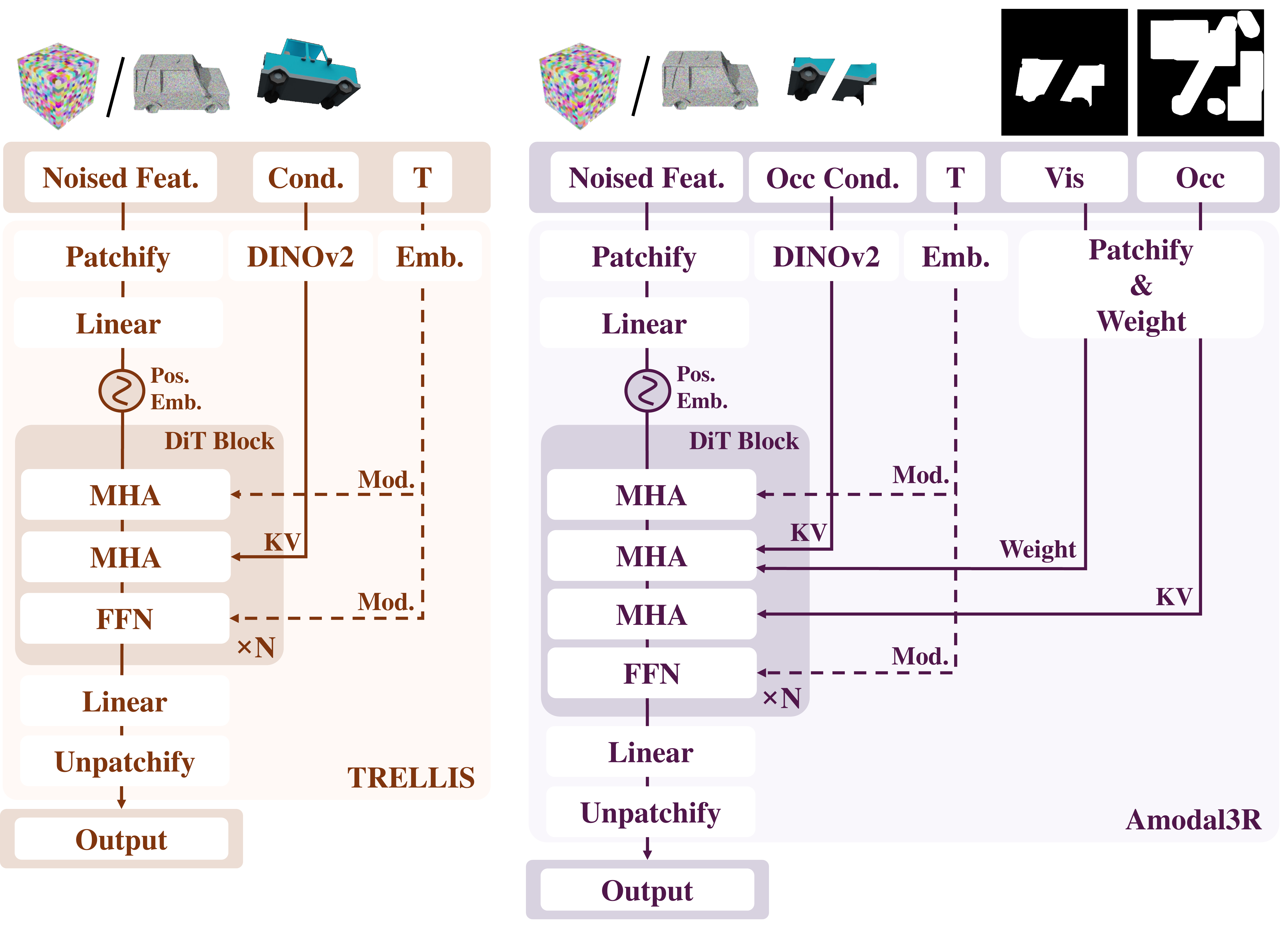}
    \begin{picture}(0,0)
      \put(-100,-3){\footnotesize \textbf{(a) TRELLIS \cite{xiang2024structured}}}
      \put(15,-3){\footnotesize \textbf{(b) \mname (Ours)}}
  \end{picture}
    \caption{\textbf{The transformer structure of \mname.} Compared with the original TRELLIS \cite{xiang2024structured} design, we further introduce the mask-weighted cross-attention and occlusion-aware layer. It applies to both sparse structure and SLAT diffusion models.
    }%
    \label{fig:block}
\end{figure}

TRELLIS comprises two diffusion models, one to predict the active voxel centers
$\{\bm{p}_i\}_{i=1}^L$
(stage 1) and the other to recover the corresponding latents
$\{\bm{z}_i\}_{i=1}^L$
(stage 2).
Each model can be viewed as a denoising neural network $\upsilon_\theta$ operating in a latent space $\bm{\ell}$, and trained to remove Gaussian noise
$
\epsilon\sim\mathcal{N}(0,\bm{I})
$
added to the latent code, \ie,
$\bm{\ell}^{(t)}=(1-t)\bm{\ell}^{(0)}+t\bm{\epsilon}$,
where $t\in[0,1]$ is the noise level~\cite{liu2022flow}.
The denoising network 
$\upsilon_\theta$ is trained to minimize the flow loss:
\begin{equation}
    \label{eq:diff}
    \min_\theta\mathbb{E}_{(\bm{\ell}^{(0)},x), t,\bm{\epsilon}}
    \|
    \upsilon_\theta(\bm{\ell}^{(t)}, x,t)
    - (\bm{\epsilon}-\bm{\ell}^{(0)})
    \|^2_2,
\end{equation}
where $x$ is the image prompt.
In stage 1, the latent code is a compressed version of the $N\times N\times N$ occupancy volume, where the spatial resolution is reduced from $N=64$ to $r=16$.
Hence, in this case the latent vector is a matrix $\bm{\ell}\in\mathbb{R}^{L'\times C'}$ of $L'=r^3=4096$ $C'$-dimensional tokens.
In stage 2, the latent code $\bm{\ell} = \{\bm{z}_i\}_{i=1}^L \in \mathbb{R}^{L\times C}$ is a matrix of $L$ $C$-dimensional tokens, where $L$ is now the number of active voxels.
Similar transformers are implemented to the corresponding denoising networks $\upsilon_\theta$ (\cref{fig:block}).
The conditioning image $x$ is read via cross-attention layers that pool information from the tokens $\bm{c}_\text{dino}$ extracted by DINO~\cite{oquabdinov2} from the image $x$.

\subsection{Mask-Conditional Generative Models}%
\label{sec:mask-weighted-cross-attn}

The key change needed to fine-tune the generator to work with partially occluded images is to condition the transformers $\upsilon_\theta$ not only on the image $x$, but also on the masks $M_\text{vis}$ and $M_\text{occ}$.
A naive approach is to downsample or embed the masks to obtain tokens $(\bm{c}_\text{vis}, \bm{c}_\text{occ})$ that can be concatenated to the image tokens $\bm{c}_\text{dino}$, to be processed by cross-attention by the transformer as before.
However, the image $x$ contains significantly more information than the binary masks $M_\text{vis}$ and $M_\text{occ}$, so the transformer, which is initially trained to consider $x$ only, may simply ignore this information.
This is compounded by the fact that learning to use this new information, which involves a new type of tokens that are incompatible with image ones, may require aggressive fine-tuning of the transformer.

To sidestep this problem,
inspired by masked attention approaches in 2D completion~\cite{zheng2022bridging},
we introduce \emph{mask-weighted cross-attention} and an \emph{occlusion-aware attention} layer to better utilize the visibility mask $M_\text{vis}$ and the occlusion mask $M_\text{occ}$ without disrupting the pre-trained 3D generator too much.
These are described next.

\paragraph{Mask-weighted Cross-Attention.}

A key novel component of our model is \emph{mask-weighted cross-attention},
which allows the model to focus its attention on the visible parts of the object.
We implement this mechanism in each attention block in the transformers $\upsilon_\theta$ of \cref{sec:preliminaries}.
Given the latent tokens $\bm{\ell}\in \mathbb{R}^{L \times C}$ input to a cross-attention layer as well as  the image features $\bm{c}_\text{dino}\in \mathbb{R}^{K\times C'}$ of the partially visible object, cross-attention computes the similarity score matrix
\begin{equation}
    \label{eq:orig_atten}
    \bm{q}={W}_q\bm{\ell},
    \quad[\bm{k},\bm{v}]
    ={W}_{kv}\bm{c}_\text{dino},
    \quad S =
    \bm{q}\bm{k}^{\top}/\sqrt{D},
\end{equation}
where ${W}_q$ and ${W}_{kv}$ are learnable projections that map the latents $\bm{\ell}$ to the query $\bm{q}$ and the conditioning image feature $\bm{c}_\text{dino}$ to the key $\bm{k}$ and the value $\bm{v}$, respectively, and $D$ is the dimension of the query and key vectors.

We wish to bias the token similarity matrix $S\in\mathbb{R}^{L\times K}$ towards the visible parts of the object.
Recall that $K$ is the number of tokens $\bm{c}_\text{dino}$ extracted by DINOv2~\cite{oquab2023dinov2} from the occluded image $x\cdot M_\text{vis}$.
Each token thus corresponds to a $P\times P$ patch in the input image (where $P = 14$).
We extract analogous flatten 1-D tokens $\bm{c}_\text{vis} = [\rho(M_\text{vis,1}),\dots,\rho(M_{\text{vis},K})]$ 
from the visibility mask by calculating the fraction of $P\times P$ pixels that are visible in the $j$-th image patch  $M_{\text{vis},j}$.
$\bm{c}_{\text{vis},j}=\rho(M_{\text{vis},j})\in[0,1]$. 
We then use these quantities to bias the computation of the attention matrix $A = \operatorname{softmax}({S}+\log c_\text{vis})\in [0,1]^{L\times M}$ via broadcasting.
Hence:
\begin{align}
\label{eq:attn_w}
&{A}_{ij}
=
\frac{\bm{c}_{\text{vis},j}\exp({S}_{ij})}
{\sum_{k=1}^K \bm{c}_{\text{vis},k}\exp({S}_{ik})}.
\end{align}
In this manner, the visibility flag modulates the attention matrix ${A}$ smoothly.
If there are no visible pixels in a patch $j$, then $A_{ij}=0$, so the corresponding image tokens are skipped in cross attention.
While we have illustrated how this works for a single head, in practice \mname uses a multiple-head transformer, to which \cref{eq:attn_w} extends trivially.
Please see the supplementary materials for details.

\begin{figure}[tb!]
    \centering
    \includegraphics[width=\linewidth]{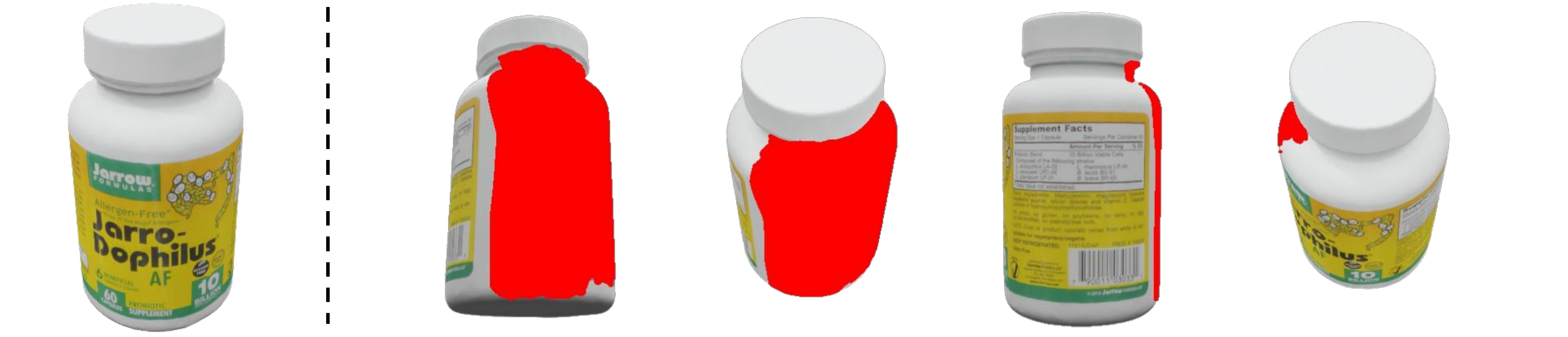}
    \vspace{-14pt}
    \caption{\textbf{3D-consistent mask example.}
    Given a 3D mesh, we render \emph{selected triangles} in a distinct color from the others to generate multi-view consistent masks. 
    It allows the evaluation of multi-view methods in handling contact occlusion. (The occluded regions are shown in \textcolor[RGB]{255,0,0}{red}.)
    }
    \label{fig:3d_mask}
\end{figure}

\paragraph{Occlusion-aware Attention Layer.}

In addition to encouraging the network to focus its attention on the visible part of the object, we also introduce an occlusion-aware attention layer.
For amodal completion, it is not enough to specify the visible information;
instead, we must also \emph{differentiate foreground occluders from the background},
as this explicitly identifies the potential regions requiring completion.
Namely, if a pixel is denoted as invisible in the mask $M_\text{vis}$, this might be because there is an occluder in front of that pixel (so the pixel \emph{could have} contained the object except due to occlusion), or because the pixel is entirely \emph{off} the object.
This information is encoded by the mask $M_\text{occ}$.

To allow the model to distinguish between visible, occluded and background areas, we add one more cross-attention layer placed immediately after the mask-weighted cross-attention layer, and pooling solely the occlusion mask $M_\text{occ}$.
In order to do so, we encode the occlusions mask $M_\text{vis}$ as a set of flatten 1-D tokens $\bm{c}_\text{occ} = [\rho(M_\text{occ,1}),\dots,\rho(M_{\text{occ},M})]$, as before and then simply pool $\bm{c}_\text{occ}$ using a cross-attention layer.

\subsection{Simulating Occluded 3D Data}%
\label{sec:mask-data}

To train and evaluate our model, we require a dataset of 3D assets imaged in scenarios with clutter and occlusions.
It is challenging to obtain such data in the real world, so we resorted to synthetic data and simulation to train our model.

\paragraph{Random 2D Occlusions.}

In order to train our model, each training sample $(\bm{o},x,M_\text{vis},M_\text{occ})$ consists of a 3D object $\bm{o}$ (from which ground truth latents can be obtained by using the encoders of~\cref{sec:preliminaries}), an image $x$ with partial occlusions, and corresponding visibility and occlusion masks $M_{\text{vis}}$ and $M_{\text{occ}}$.
In a real scenario, occlusions arise from other objects in the scene.
In a multi-view setting, like the one discussed below, these occlusions need to be consistent across views, reflecting the underlying scene geometry.
However, because our model is trained for single-view reconstruction, we can simulate occlusions by randomly masking parts of the object after rendering it in 2D.
This is simpler and allows us to generate fresh occlusion patterns each time a view is sampled for training.

Thus, given $\bm{o}$ and an image $x$ rendered from a random viewpoint, we generate random occlusion masks as follows.
Inspired by work on 2D completion~\cite{pathak2016context,iizuka2017globally,Liu_2018_ECCV,zheng2019pluralistic}, we randomly place lines, ellipses, and rectangles, simulating diverse masking patterns.
The union of these random shapes gives us the occlusion mask $M_{\text{occ}}$.
Then, if $M_\text{obj}$ is the mask of the unoccluded object, the visible mask is given by $M_\text{vis} = M_\text{obj} \odot (1 - M_\text{occ})$.
Examples of such patterns are given in the supplementary material.

\paragraph{3D-consistent occlusions.}

In a real scene imaged from multiple viewpoints, occlusions are not random but consistent across views as they are caused by other objects.
This is particularly true for \emph{contact occlusions}, where part of an object remains occluded by another from \emph{all} viewpoints.
To evaluate the model's performance under such challenging conditions, 3D-consistent masks are required.

To efficiently generate such contact occlusion masks, we leverage 3D meshes during rendering.
Starting from a randomly chosen triangle, we apply a random-walk strategy to iteratively select neighboring triangles, forming continuous occluded regions until the predefined mask ratio is met.
By rendering these masked meshes using the same camera parameters as the RGB images, we ensure multi-view consistency in the generated occlusion masks (see~\cref{fig:3d_mask}).

\subsection{Reconstruction with Multi-view Input}

Since our flow-based model performs multiple denoising steps and does \emph{not} require known camera poses for input views,
\mname can accept \emph{multi-view} reference images $\mathcal{X}=\{x_i\}_{i=1}^{N}$ as conditioning inputs at different steps of the denoising process.
While multi-view conditioning naturally improves reconstruction performance, a potential issue with such multi-view conditioning is that an image used earlier in the denoising process has a stronger influence on the final 3D geometry.
This is because early denoising steps establish the coarse geometry, whereas later steps refine higher-frequency details~\cite{kim2022diffusionclip,kim2023leveraging}.
Therefore, we prioritize input images based on their visibility.
Specifically, when experimenting with multi-view inputs, we sort the images in proportion to the object visibility $|M_{\text{vis}}|$, ensuring that images with higher visibility are used as primary inputs.
\begin{table*}[tb!]
  \centering
  \resizebox{\linewidth}{!}{
  \begin{tabular}{@{}lccccccccc@{}}
    \toprule
    Method & V-num & 2D-Comp & $\text{FID}\downarrow$ & $\text{KID}(\%)\downarrow$ & $\text{CLIP}\uparrow$ & $\text{P-FID}\downarrow$ & $\text{COV}(\%)\uparrow$ & $\text{MMD}(\text{\textperthousand})\downarrow$\\
    \midrule
    GaussianAnything \cite{lan2024gaussiananything} & 1 & pix2gestalt \cite{ozguroglu2024pix2gestalt} & 92.26 & \cellcolor{orange!30}1.30 & 0.74 & 34.69 & \cellcolor{orange!30}35.92 & 5.03 \\
    Real3D \cite{jiang2024real3d} & 1 & pix2gestalt \cite{ozguroglu2024pix2gestalt} & 91.21 & 2.02 & 0.75 & \cellcolor{orange!30}23.92 & 19.61 & 9.21\\
    TRELLIS \cite{xiang2024structured} & 1 & pix2gestalt \cite{ozguroglu2024pix2gestalt} & \cellcolor{orange!30}58.82 & 5.87 & \cellcolor{orange!30}0.76  & 26.43 & 31.65 & \cellcolor{orange!30}4.17 \\
    \textbf{\mname (Ours)} & 1 & - & \cellcolor{red!30}\textbf{30.64} & \cellcolor{red!30}\textbf{0.35} & \cellcolor{red!30}\textbf{0.81}  & \cellcolor{red!30}\textbf{7.69} & \cellcolor{red!30}\textbf{39.61} & \cellcolor{red!30}\textbf{3.62} \\
    \midrule
    LaRa \cite{chen2024lara} & 4 & pix2gestalt \cite{ozguroglu2024pix2gestalt} & 172.84 & 4.54 & 0.70 & 66.34  & 24.56 & 8.11 \\
    LaRa \cite{chen2024lara} & 4 & pix2gestalt \cite{ozguroglu2024pix2gestalt}+MV & 97.53 & 2.63 & 0.75 & 21.80 & 26.21 & 8.61 \\
    TRELLIS \cite{xiang2024structured} & 4 & pix2gestalt \cite{ozguroglu2024pix2gestalt} & 65.69 & 6.92  & 0.78 & 24.64 & \cellcolor{orange!30}32.33 & 4.26  \\
    TRELLIS \cite{xiang2024structured} & 4 & pix2gestalt \cite{ozguroglu2024pix2gestalt}+MV & \cellcolor{orange!30}60.37 & \cellcolor{orange!30}1.85 & \cellcolor{orange!30}0.83 & \cellcolor{orange!30}19.68 & 31.75 & \cellcolor{orange!30}4.21 \\
    \textbf{\mname (Ours)} & 4 & - & \cellcolor{red!30}\textbf{26.27} & \cellcolor{red!30}\textbf{0.22} & \cellcolor{red!30}\textbf{0.84} & \cellcolor{red!30}\textbf{5.03} & \cellcolor{red!30}\textbf{38.74} & \cellcolor{red!30}\textbf{3.61} \\
    \bottomrule
  \end{tabular}}
  \vspace{-6pt}
  \caption{\textbf{Amodal 3D reconstruction results on GSO~\cite{downs2022google}.}
  Here, V-num denotes the number of input views, and 2D-Comp refers to the 2D completion method.
  For single-view image-to-3D, we first complete occluded objects using pix2gestalt \cite{ozguroglu2024pix2gestalt} before passing them to various 3D models.
  For multi-view image-to-3D, we adopt two variants:
  1) pix2gestalt \cite{ozguroglu2024pix2gestalt} is applied independently on each view;
  2) pix2gestalt \cite{ozguroglu2024pix2gestalt} + MV: a single-view completion is generated first, followed by multi-view diffusion~\cite{shi2023zero123++} to synthesize 4 views as inputs.
  Without bells and whistles, \mname consistently outperforms state-of-the-art models across all evaluation metrics.
  }
  \label{tab: gso}
\end{table*}
\begin{table*}[tb!]
  \centering
  \resizebox{\linewidth}{!}{
  \begin{tabular}{@{}lccccccccc@{}}
    \toprule
    Method & V-num & 2D-Comp & $\text{FID}\downarrow$ & $\text{KID}(\%)\downarrow$ & $\text{CLIP}\uparrow$ & $\text{P-FID}\downarrow$ & $\text{COV}(\%)\uparrow$ & $\text{MMD}(\text{\textperthousand})\downarrow$\\
    \midrule
    GaussianAnything \cite{lan2024gaussiananything} & 1 & pix2gestalt \cite{ozguroglu2024pix2gestalt} & 57.17 & \cellcolor{orange!30}1.22  & 0.80  & \cellcolor{orange!30}21.97 & \cellcolor{orange!30}33.56 & 7.23 \\
    Real3D \cite{jiang2024real3d} & 1 & pix2gestalt \cite{ozguroglu2024pix2gestalt} & 59.92 & 1.63 & 0.79 & 23.31 & 24.35 & 9.53\\
    TRELLIS \cite{xiang2024structured} & 1 & pix2gestalt \cite{ozguroglu2024pix2gestalt} & \cellcolor{orange!30}43.05 & 6.83 & \cellcolor{orange!30}0.80 & 26.04 & 26.28 & \cellcolor{orange!30}6.87 \\
    \textbf{\mname (Ours)} & 1 & - & \cellcolor{red!30}\textbf{23.45} & \cellcolor{red!30}\textbf{0.42} & \cellcolor{red!30}\textbf{0.83} &  \cellcolor{red!30}\textbf{5.00} & \cellcolor{red!30}\textbf{37.09} & \cellcolor{red!30}\textbf{5.89}\\
    \midrule
    LaRa \cite{chen2024lara} & 4 & pix2gestalt \cite{ozguroglu2024pix2gestalt} & 123.52 & \cellcolor{orange!30}3.61 & 0.75 & 45.91 & \cellcolor{orange!30}27.89 & 9.67 \\
    LaRa \cite{chen2024lara} & 4 & pix2gestalt \cite{ozguroglu2024pix2gestalt}+MV & 75.33 & 4.14 & 0.80 & \cellcolor{orange!30}13.00 & 24.82 & 10.93 \\
    TRELLIS \cite{xiang2024structured} & 4 & pix2gestalt \cite{ozguroglu2024pix2gestalt} & 46.34 & 8.77 & 0.81 & 28.76 & 25.35 & 7.13 \\
    TRELLIS \cite{xiang2024structured} & 4 & pix2gestalt \cite{ozguroglu2024pix2gestalt}+MV & \cellcolor{orange!30}43.00 & 7.53 & \cellcolor{orange!30}0.81 & 24.41 & 26.55 & \cellcolor{orange!30}7.05 \\
    \textbf{\mname (Ours)} & 4 & - & \cellcolor{red!30}\textbf{20.93} & \cellcolor{red!30}\textbf{0.50} & \cellcolor{red!30}\textbf{0.85} & \cellcolor{red!30}\textbf{3.78} & \cellcolor{red!30}\textbf{39.03} & \cellcolor{red!30}\textbf{5.75}\\
    \bottomrule
  \end{tabular}}
  \vspace{-6pt}
  \caption{\textbf{Amodal 3D reconstruction results on Toys4K~\cite{stojanov2021using}.}
  The experimental setting is the same to~\cref{tab: gso}, except for the dataset.
  }
  \label{tab: toys4k}
\end{table*}

\section{Experiments}%
\label{sec:exp}

\subsection{Experiment Settings}

\paragraph{Datasets.}

\mname is trained on a combination of 3D synthetic dataset:
3D-FUTURE (9,472 objects~\cite{fu20213d}), ABO (4,485 objects~\cite{collins2022abo}), and HSSD (6,670 objects~\cite{khanna2024habitat}).
Once \mname is trained, we first assess its effectiveness on Toys4K (randomly sampling 1,500 objects~\cite{stojanov2021using}) and Google Scanned Objects (GSO) (1,030 objects~\cite{downs2022google}), which are excluded from our training data for our and the baseline model.
During inference,
a 3D-consistent mask occludes each input object, and each view is augmented with additional randomly generated occlusion areas.
This ensures that the model cannot directly extract the region required to be completed from the occlusion regions.
To further assess the \emph{out-of-distribution} generalization ability in practical applications, we also evaluate all models on the 3D scene dataset Replica~\cite{straub2019replica} as well as on in-the-wild images.
The training and evaluation dataset will be released.

\paragraph{Metrics.}

To measure the quality of the rendered images, we use the Fr$\Acute{e}$chet Inception Distance (FID)~\cite{heusel2017gans} and the Kernel Inception Distance (KID)~\cite{binkowski2018demystifying}.
To measure the quality of the reconstructed 3D geometry, we use the Coverage Score (COV), the Point cloud FID (P-FID)~\cite{nichol2022point}, and the Minimum Matching Distance (MMD) using the Chamfer Distance (CD).
COV measures the diversity of the results and P-FID and MMD the quality of the 3D reconstruction.
We also use the CLIP score~\cite{radford2021learning} to evaluate the consistency between each pair generated and ground-truth objects.

\paragraph{Baselines.}

Most 3D generative models are trained on complete object inputs.
To ensure fair comparisons, we complete the partially visible 2D objects before passing them to 3D generators.
Sepecifically, we use pix2gestalt~\cite{ozguroglu2024pix2gestalt}, a state-of-the-art 2D amodal completion network.
Using this protocol, we compare \mname to state-of-the-art methods such as TRELLIS~\cite{xiang2024structured}, Real3D~\cite{jiang2024real3d}, GaussianAnything~\cite{lan2024gaussiananything}$_\text{ICLR'25}$, and LaRa~\cite{chen2024lara}$_\text{ECCV'24}$.

\paragraph{Implementation Details.}

\mname is trained on 4 A100 GPUs (40G) for 20K steps with a batch size of 16, taking approximately a day.
Following TRELLIS~\cite{xiang2024structured}, we implement classifier-free guidance (CFG~\cite{ho2022classifier}) with a drop rate of 0.1 and AdamW~\cite{loshchilov2017decoupled} optimizer with a learning rate of 1e-4.
More details are provided in the supplementary material.

\begin{figure*}[tb!]
  \centering
  \includegraphics[width=\textwidth]{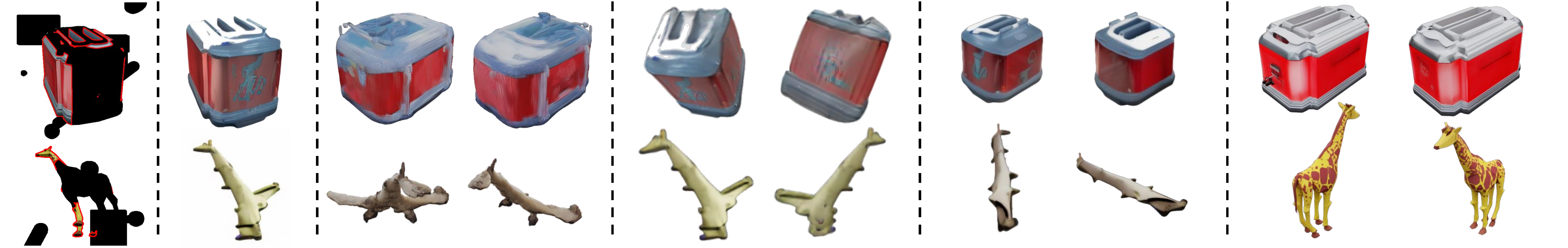}
  \begin{picture}(0,0)
      \put(-225,3){\footnotesize Input}
      \put(-195,3){\footnotesize pix2gestalt \cite{ozguroglu2024pix2gestalt}}
      \put(-135,3){\footnotesize GaussianAnything \cite{lan2024gaussiananything}}
      \put(-22,3){\footnotesize Real3D \cite{jiang2024real3d}}
      \put(72,3){\footnotesize TRELLIS \cite{xiang2024structured}}
      \put(160,3){\footnotesize \mname (Ours)}
  \end{picture}
  \vspace{-6pt}
  \captionof{figure}{\textbf{Single-view amodal 3D reconstruction.} The occlusion regions are shown in black and the visible regions are highlighted with red outlines. More examples are provided in supplementary material Fig.~C.4.}
  \label{fig:single-view}
\end{figure*}
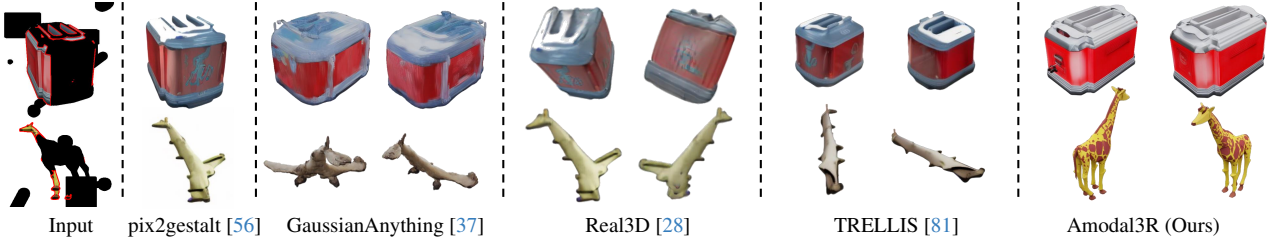
\begin{figure*}
  \centering
  \includegraphics[width=\textwidth]{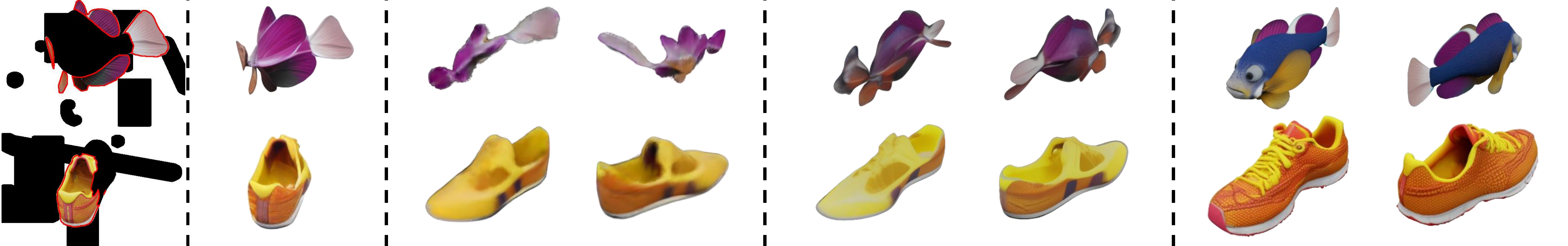}
  \begin{picture}(0,0)
      \put(-225,3){\footnotesize Input}
      \put(-180,3){\footnotesize pix2gestalt \cite{ozguroglu2024pix2gestalt}}
      \put(-75,3){\footnotesize LaRa \cite{chen2024lara}}
      \put(40,3){\footnotesize TRELLIS \cite{xiang2024structured}}
      \put(160,3){\footnotesize \mname (Ours)}
  \end{picture}
  \vspace{-6pt}
  \captionof{figure}{\textbf{Multi-view amodal 3D reconstruction.}
  The above results are reconstructed using 4 occluded input views.
  Due to limited space, we present only the best results for LaRa and TRELLIS under the "pix2gestalt+MV" setting.
  We apply 3d-consistent mask and random extended occlusion areas.
  More examples are provided in supplementary material Fig.~C.5.
  }
  \label{fig:multi-view}
\end{figure*}
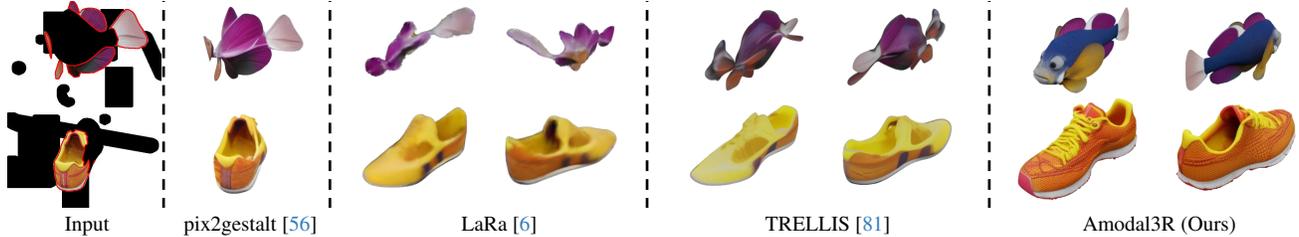
\subsection{Main Results}

\paragraph{Quantitative Results.}

We compare \mname to the state-of-the-art for \emph{amodal 3D reconstruction} in~\cref{tab: gso,tab: toys4k}.
\mname significantly outperforms the baselines across all metrics in both single- and multi-view 3D reconstruction, demonstrating its effectiveness.
Interestingly, baselines that use multiple but potentially inconsistently-completed views (``4 V-num + pix2gestalt'') are worse than using a single completed view (``1 V-num + pix2gestalt'').
This shows that \emph{inconsistent 2D completion} does confuse reconstruction models to the point that using a single view is preferable (\cref{sec:background}).
This issue does not affect \mname as it does not rely on 2D completion; in our case, utilizing more occluded views does improve the results as expected.

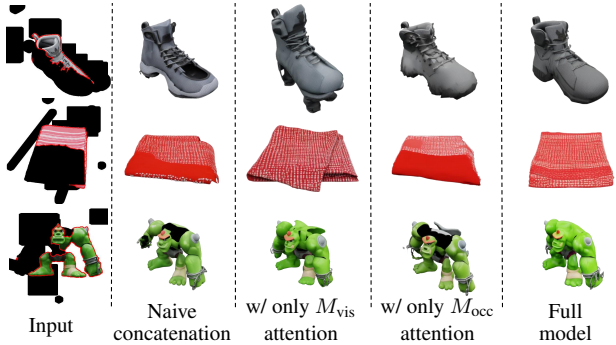
\begin{figure}[tb!]
    \centering
    \includegraphics[width=\linewidth]{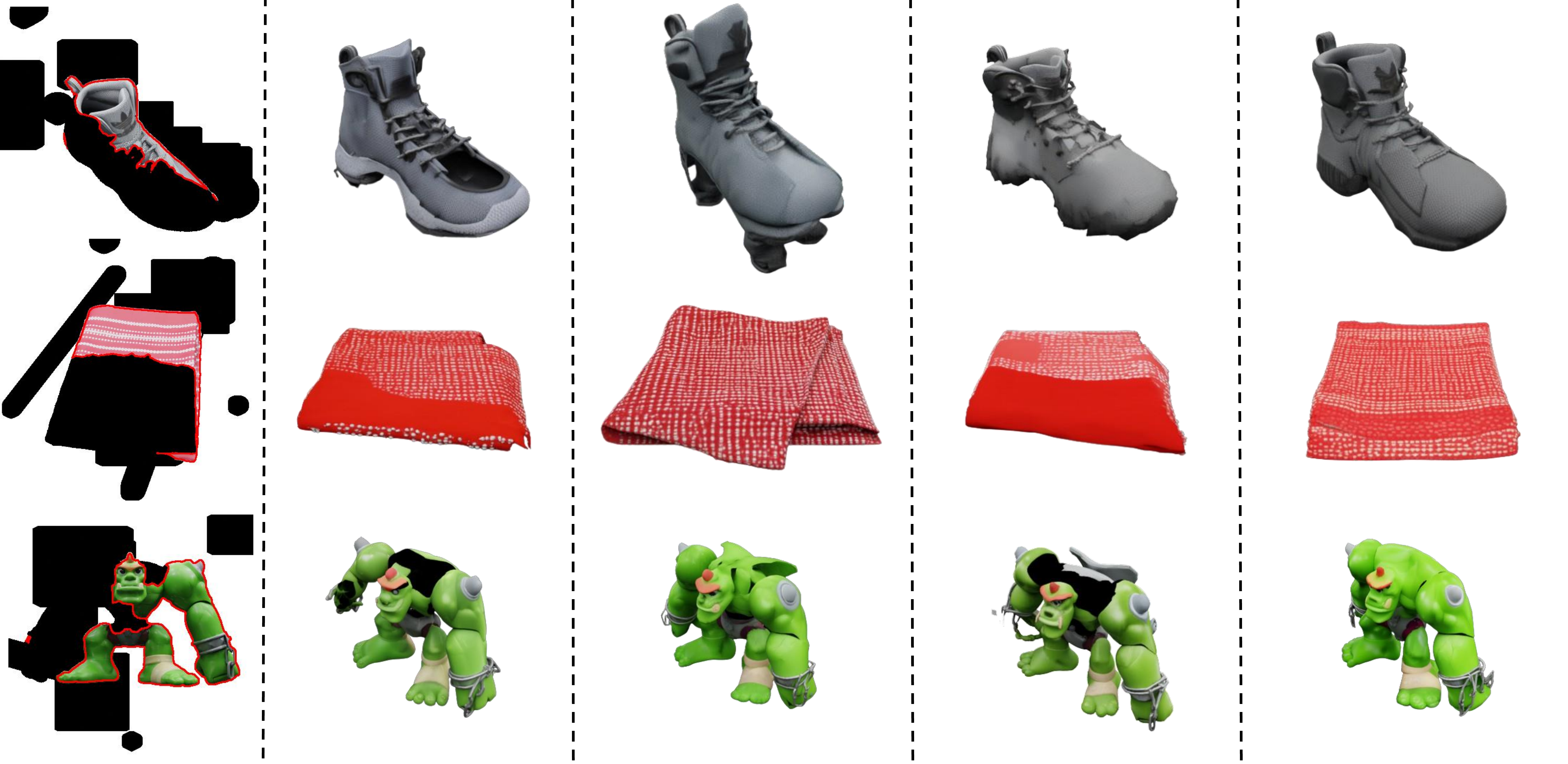}
    \begin{picture}(0,0)
      \put(-110,3){\footnotesize Input}
      \put(-78,0){\footnotesize \shortstack{Naive \\concatenation}}
      \put(-28,0){\footnotesize \shortstack{w/ only $M_\text{vis}$ \\attention}}
      \put(23,0){\footnotesize \shortstack{w/ only $M_\text{occ}$ \\attention}}
      \put(83,0){\footnotesize \shortstack{Full\\model}}
  \end{picture}
    \caption{\textbf{Ablation examples.}
    Naive concatenation fails to reconstruct complete shape and appearance.
    Mask-weighted attention alone extends geometry into background regions, while occlusion-aware attention alone cannot guarantee photorealistic appearance.
    }
    \label{fig:ablation}
\end{figure}

\paragraph{Qualitative Results.}

The qualitative results are shown in~\cref{fig:single-view,fig:multi-view} and in supplementary material Figs.~C.4 and C.5.
For all baselines, pix2gestalt is first applied for 2D amodal completion (second column), and the completed images are passed to each baseline image-to-3D model.
\mname produces high-quality 3D assets even under challenging conditions in both single-view and multi-view image-to-3D tasks.
In contrast, 2D amodal completions inconsistencies accumulate as more views are added, particularly when pix2gestalt is more uncertain, confusing the reconstructor models downstream.
For instance, in the giraffe example in~\cref{fig:single-view}, the pix2gestalt completion fails to capture the overall structure of the 3D object correctly, which in turn leads the 3D generator models to misinterpret it as a woodstick-like shape.
In contrast, \mname reconstructs the 3D geometry and appearance well, with good alignment to the occluded inputs.
These findings highlight not only the effectiveness of \mname
but also the advantage of completing objects while reconstructing them, which avoids relying on monocular 2D completion models that may introduce inconsistencies.

\begin{table}[tb!]
  \centering
  \setlength\tabcolsep{3pt}
  \resizebox{\linewidth}{!}{  
  \begin{tabular}{@{}lcccc@{}}
    \toprule
    Method &  $\text{FID}\downarrow$ & $\text{KID}(\%)\downarrow$ & $\text{COV}(\%)\uparrow$ & $\text{MMD}(\text{\textperthousand})\downarrow$\\
    \midrule
    naive conditioning & 31.96 & 0.49 & 37.96 & \underline{3.61}  \\
    w/ only mask-weighted attention & \textbf{30.53} & \underline{0.38} & 36.90 & 3.69  \\
    w/ only occlusion-aware layer & 31.77 & 0.57 & \textbf{40.19} & \textbf{3.51}  \\
    full model (\textbf{Ours}) & \underline{30.64} & \textbf{0.35} & \underline{39.61} & 3.62 \\
    \bottomrule
  \end{tabular}}
  \vspace{-6pt}
  \caption{\textbf{Ablations on different mask conditioning designs.}
  The \textbf{Best} and the \underline{second best} results are highlighted.
  }
  \label{tab:ablation}
\end{table}
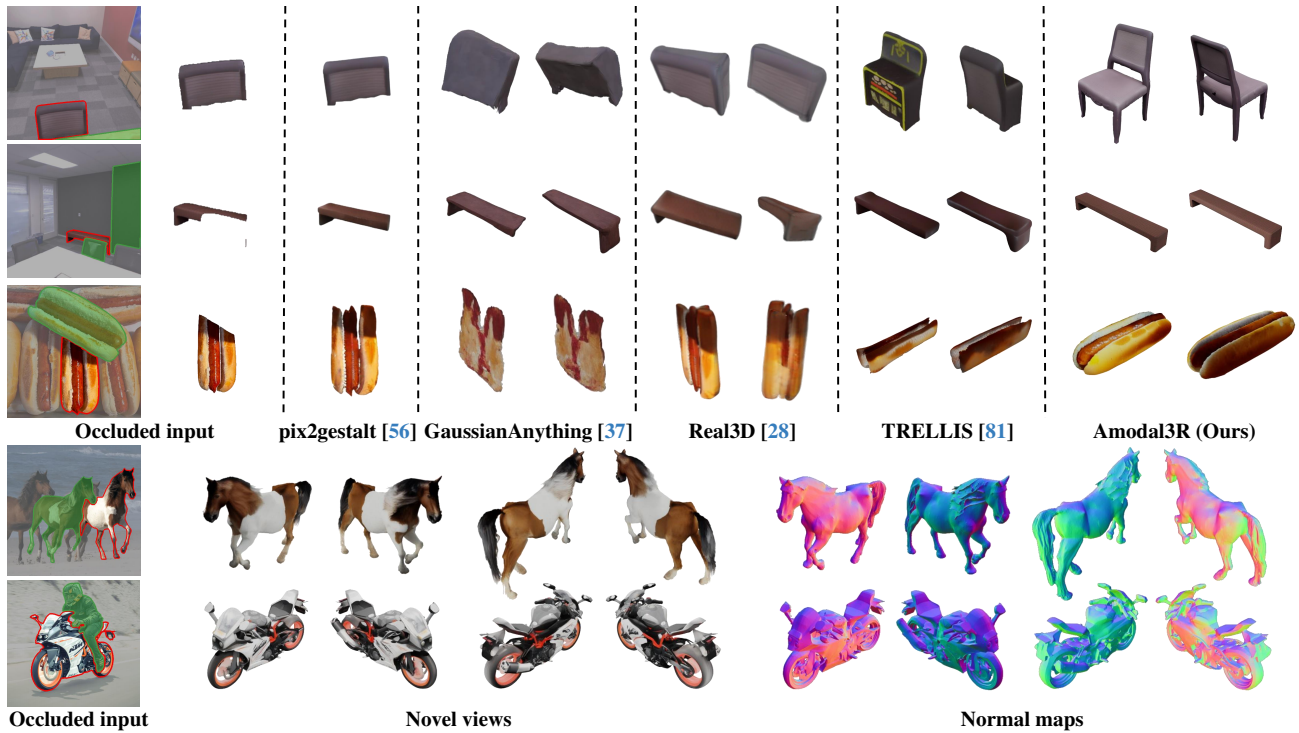
\begin{figure*}[tb!]
    \centering
    \includegraphics[width=\linewidth]{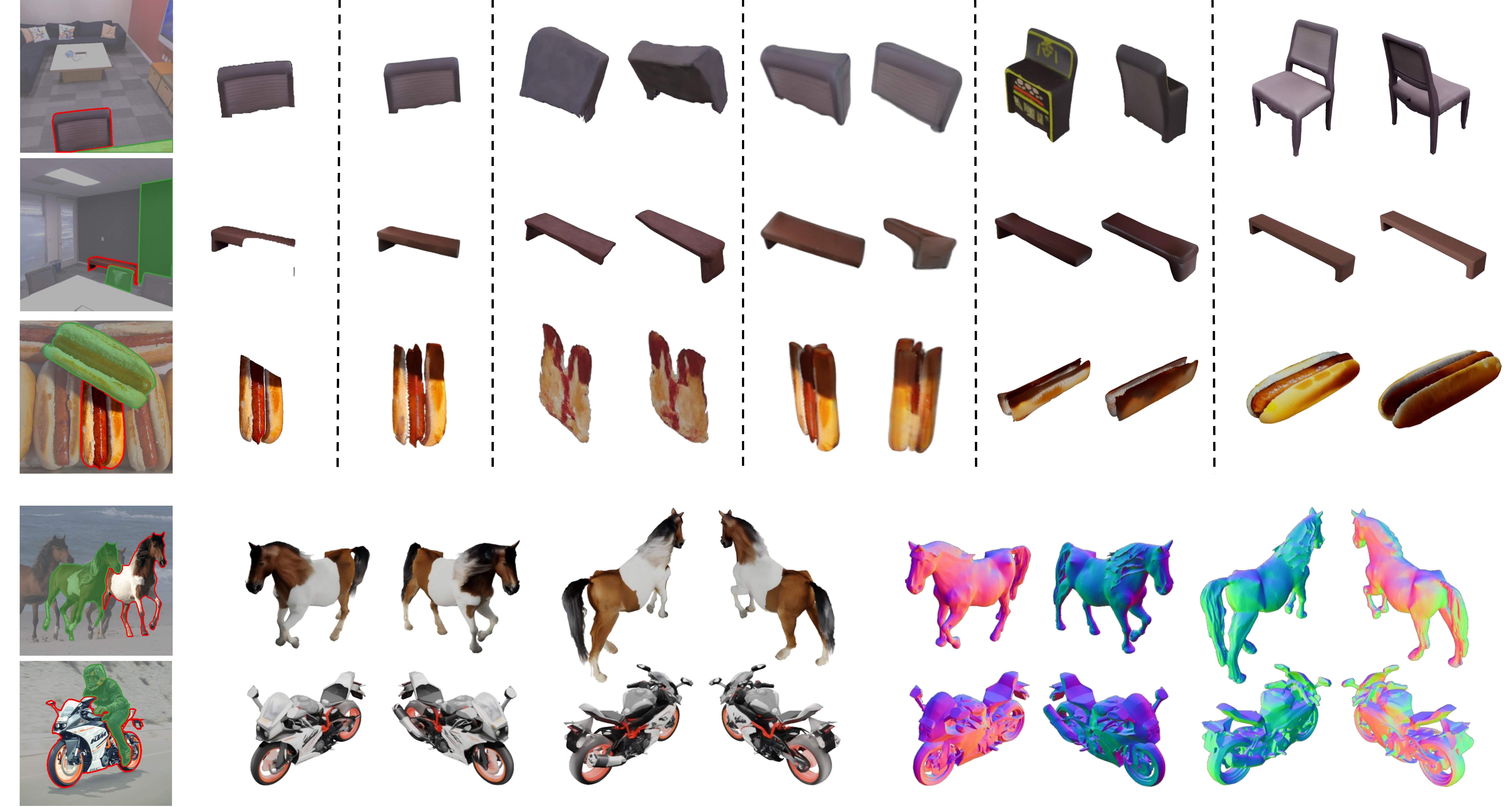}
    \begin{picture}(0,0)
      \put(-215,114){\footnotesize \textbf{Occluded input}}
      \put(-138,114){\footnotesize \textbf{pix2gestalt \cite{ozguroglu2024pix2gestalt}}}
      \put(-83,114){\footnotesize \textbf{GaussianAnything \cite{lan2024gaussiananything}}}
      \put(17,114){\footnotesize \textbf{Real3D \cite{jiang2024real3d}}}
      \put(90,114){\footnotesize \textbf{TRELLIS \cite{xiang2024structured}}}
      \put(170,114){\footnotesize \textbf{\mname (Ours)}}
      \put(-240,5){\footnotesize \textbf{Occluded input}}
      \put(-90,5){\footnotesize \textbf{Novel views}}
      \put(120,5){\footnotesize \textbf{Normal maps}}
  \end{picture}
  \vspace{-8pt}
    \caption{\textbf{Examples on Replica \cite{straub2019replica} and in-the-wild images.} The target objects and the occluders are shown in \textcolor[RGB]{250,0,0}{red} and \textcolor[RGB]{68,153,69}{green} outlines.}
    \label{fig:realcase}
\end{figure*}

\subsection{Ablation Study}

We conducted several ablations to study the impact of the different components of \mname and report the results in~\cref{tab:ablation,fig:ablation}.
For these experiments, we test single-view image-to-3D reconstruction on the GSO dataset.

\paragraph{Naive Conditioning.}

We first evaluated a version of the model that still conditions the reconstruction on the visibility and occlusion masks, but without using the modules of \cref{sec:mask-weighted-cross-attn}.
Instead, we directly concatenate the tokens $\bm{c}_\text{vis}$  
to $\bm{c}_\text{dino}$ to form the condition for the cross-attention layer.
The results (first row in \cref{tab:ablation} and second column in \cref{fig:ablation}) show that, while the resulting model can still perform basic completion, the textures in the occluded regions are inconsistent with those in the visible ones, and the reconstructed geometry is inaccurate, \eg the hole in the shoe.

\paragraph{Mask-weighted Attention.}

To evaluate the effectiveness of our proposed mask-weighted attention mechanism, we integrate it into the training while omitting the occlusion-aware layer.
The results demonstrate improved rendering quality --- especially in capturing texture details in the visible areas --- and significantly enhanced appearance consistency.
However, the geometries exhibit deficiencies, as seen in the problematic shoe in \cref{fig:ablation} (first row), and the toy monster with a broken back (third row).

\paragraph{Occlusion-aware Layer.}

The integration of the occlusion-aware layer improves the geometry both quantitatively and qualitatively.
This aligns with our motivation for introducing an additional cross-attention layer, aimed at reconstructing the visible areas via the image-conditioned layer and occluded areas via the subsequent layer.
However, occlusion-aware layer alone still results in unsatisfactory appearances, which again indirectly attests to the effectiveness of mask-weighted attention mechanism.

\paragraph{Full Model.}

Consequently, the full model integrating both modules achieves optimal 3D generation results characterized by complete geometry and consistent textures.

\subsection{Real-World Generation / Completion}

\mname is inherently superior in generalizing to \emph{out-of-distribution} amodal 3D reconstruction, primarily due to the fact that we build upon the model on a ``foundation'' 3D generator, and fine-tune it with diverse categories.
To demonstrate this advantage,
we conduct scene-level dataset evaluations:
Replica~\cite{straub2019replica} (first two rows) and on in-the-wild images (3rd-5th rows) in Fig.~\ref{fig:realcase}. 
Here we adopt Segment Anything~\cite{kirillov2023segment} to get the visibility and occlusion masks.
The results show that \mname generates reasonable 3D assets,
whereas pix2gestalt fails to infer complete shapes from the same inputs, leading to unsatisfactory results from subsequent image-to-3D models. We also visualize the colored normal maps, which show that the results of \mname are simultaneously rich in geometric and textural detail.

\section{Conclusion}
\label{sec:conclusion}
We have introduced \mname,
a novel approach to reconstruct complete 3D shape and appearance from partially visible 2D images.
The key to our success is that we construct mask-weighted cross-attention mechanism and occlusion-aware layer to effectively exploit visible and occluded information.
Compared to the state-of-the-art methods that rely on sequential 2D completion followed by 3D generation,
our \mname achieves remarkably better performance by operating directly in 3D space.
Furthermore, results on in-the-wild images indicate its potential for subsequent applications in 3D decomposition and scene understanding,
marking a step towards robust 3D asset reconstruction in real-world environments with complex occlusion.

{
    \small
    \bibliographystyle{ieeenat_fullname}
    \bibliography{main,chuanxia_general,chuanxia_specific}
}

\clearpage
\appendix\onecolumn
\renewcommand{\theequation}{\thesection.\arabic{equation}}
\setcounter{equation}{0}
\renewcommand{\thefigure}{\thesection.\arabic{figure}}
\setcounter{figure}{0}
\renewcommand{\thetable}{\thesection.\arabic{table}}
\setcounter{table}{0}
\setcounter{page}{1}
\maketitlesupplementary

\section{Implementation Details}
\label{sec:sup}

\subsection{Network Design}
We adopt the network design in TRELLIS \cite{xiang2024structured} to load the pre-trained image-to-3D weights and integrate the mask-weighted cross-attention mechanism to each DiT block (24 blocks in total). And each image-conditioned cross-attention layer is immediately followed by an occlusion-aware cross-attention layer.

\noindent\textbf{(a) Patchify and weight of visibility/occlusion mask} The input condition image has a resolution of 512$\times$512, which is resized to 518$\times$518 to facilitate splitting into patches of size 14$\times$14, as required by DINOv2 \cite{oquab2023dinov2}. The resulting condition is then flattened into a tensor $\textbf{c}_{dino}\in\mathbb{R}^{1374\times 1024}$, where the sequence length corresponds to 37$\times$37 patches plus 1 CLS token and 4 register tokens. To better align the visibility and occlusion masks with the DINOv2 features, we first split the masks into patches of the same size, then calculate the weight score for each patch using Eq.~4 and Eq.~5. The final $\textbf{c}_{vis}\in\mathbb{R}^{1374\times1}$ and $\textbf{c}_{occ}\in\mathbb{R}^{1374\times1}$ are obtained by flattening the weight scores, with a value of 1 assigned to the CLS and register token positions.

\noindent\textbf{(b) Occlusion-aware cross-attention layer.} We set the feature dimension of the occlusion-aware cross-attention layers to 1024, matching that of the image-conditioned cross-attention layers. To maintain consistent dimensions, we replicate the flattened occlusion masks to form a tensor $\textbf{c}_{occ\_stack}\in\mathbb{R}^{1374\times1024}$.

\noindent\textbf{(c) Multi-head Cross-Attention.} Our mask-weighted multi-head cross-attention (MHA) layer, which is implemented to encourage the model to focus its attention on the visible parts of the object, is an extension of the cross-attention described in the main paper. Specifically, $H$ heads are run in parallel, resulting in $H$ attention scores. For mask-weighted attention mechanism, we impose $\bm{c}_\text{vis}$ simultaneously to each head:

\begin{align}
    \label{eq:mh_attn_w}
    &\bm{A}_h=\operatorname{softmax}(\bm{S}_h+\log\bm{c}_\text{vis}), \\
    &\text{MHA} = [\bm{A}_1\bm{v};\bm{A}_2\bm{v};\dots;\bm{A}_H\bm{v}]
\end{align}

\subsection{Training Details}
\noindent\textbf{(a) Pre-trained model loading.} While TRELLIS is split into multiple modules, in our work we only train the sparse structure flow transformer and the structured latent flow transformer (see the overview figure where the "fire" symbols indicate the parts that are fine-tuned, and "snowflake" symbols indicate that we directly use the pre-trained weights).

\begin{figure}[ht]
    \centering
    \includegraphics[width=\linewidth]{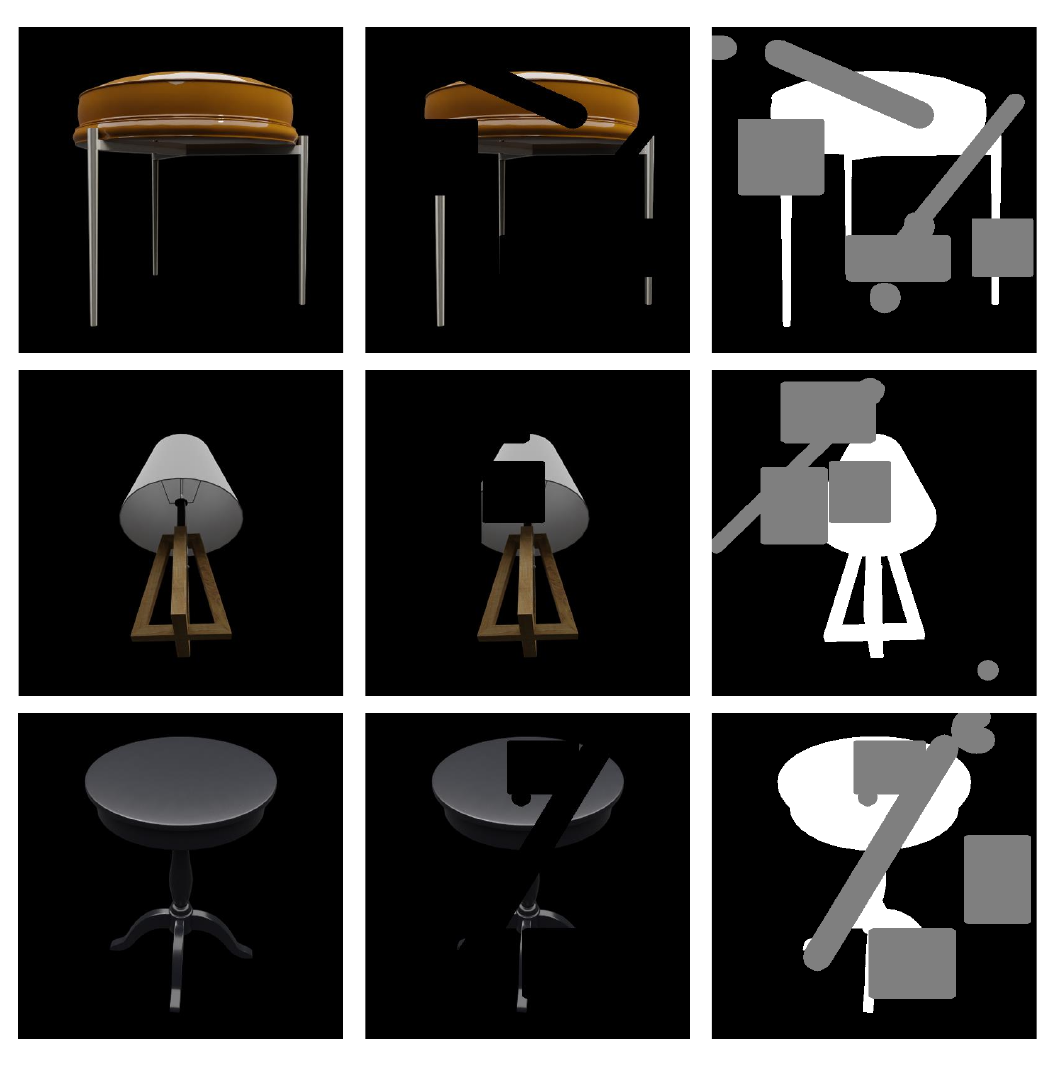}
    \begin{picture}(0,0)
      \put(-103,3){\footnotesize \textbf{Original image}}
      \put(-25,3){\footnotesize \textbf{Masked image}}
      \put(52,3){\footnotesize \textbf{Occlusion mask}}
  \end{picture}
    \caption{\textbf{Examples of random mask generation.} The visible areas are shown in white, occluded areas in gray and background in black.}
    \label{fig:training_mask_example}
\end{figure}
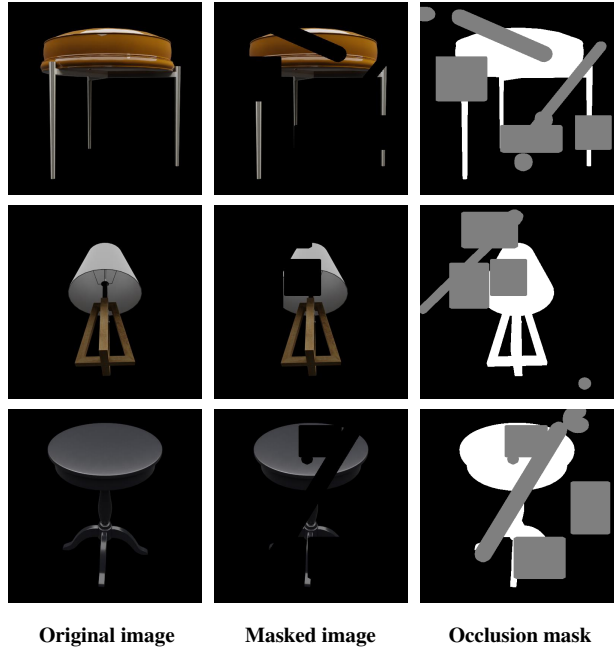

\noindent\textbf{(b) Data Augmentation.} As described in Sec.~3.3, we generate random masks during training for data augmentation. Specifically, we begin by randomly drawing 1 to 3 lines, circles, and ellipses in the mask image. Next, to ensure these regions connect --- thereby better simulating real-world occlusions, where mask regions are typically not highly fragmented --- we randomly add 3 to 7 rectangular regions that have undergone an expansion operation. This results in a stable masking of the objects in the training data. Example inputs are presented in \cref{fig:training_mask_example}.

\subsection{Inference Details}
\noindent\textbf{(a) 3D-consistent mask ratio.} For the multi-view 3D-consistent masks described in Sec.~3.3, we set the mask ratio to a random number between 0.4 and 0.6 for each object, which results in a variety of reasonable mask areas.

\noindent\textbf{(b) Time consumption.} Despite the introduction of additional cross-attention layers, our inference time remains comparable to that of the baselines. \mname can generate and render each object in under 10 seconds.

\begin{figure*}[!ht]
    \centering
    \includegraphics[width=\linewidth]{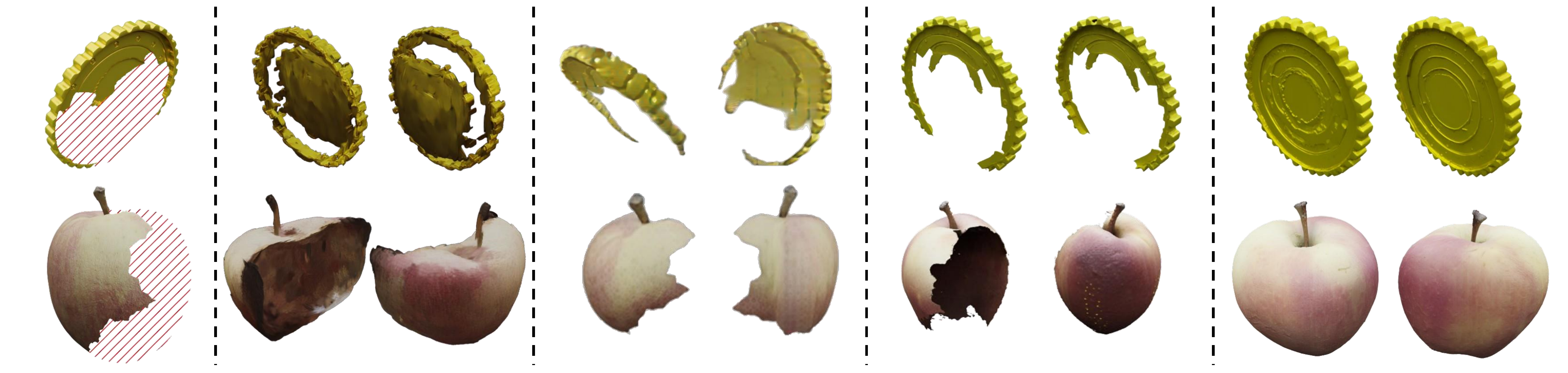}
    \begin{picture}(0,0)
      \put(-238,3){\footnotesize \textbf{Occluded input}}
      \put(-168,3){\footnotesize \textbf{GaussianAnything \cite{lan2024gaussiananything}}}
      \put(-50,3){\footnotesize \textbf{Real3D \cite{jiang2024real3d}}}
      \put(55,3){\footnotesize \textbf{TRELLIS \cite{xiang2024structured}}}
      \put(165,3){\footnotesize \textbf{\mname (Ours)}}
  \end{picture}
    \caption{\textbf{Examples using occluded images directly as the input of baseline models.}}
    \label{fig:occluded input}
\end{figure*}
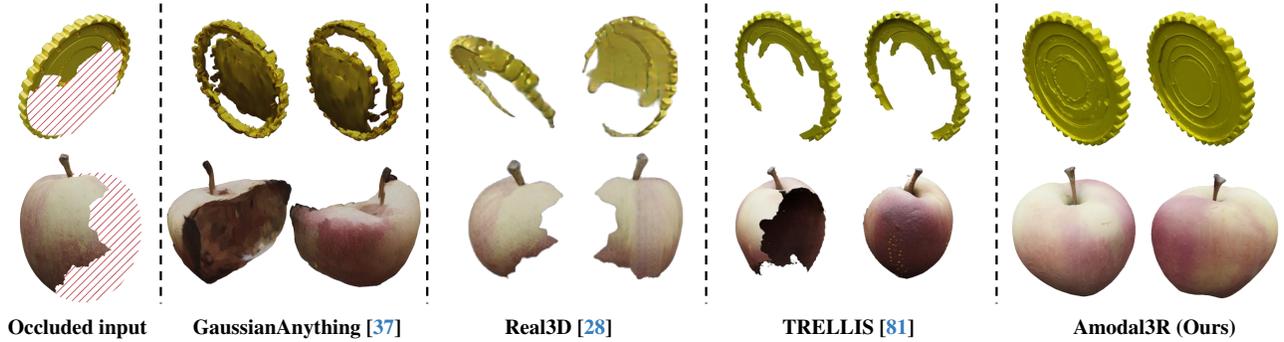

\section{Experimental Details}

\subsection{Evaluation Protocol}

We evaluate the results using Google Scanned Objects (GSO) (1,030 objects) \cite{downs2022google} and a randomly sampled subset of Toys4K \cite{stojanov2021using} containing 1,500 objects. Here, we provide additional details regarding the computation of our evaluation metrics.

\noindent\textbf{(a) Rendering quality and semantic consistency alignment}
To assess overall rendering quality, we compute the Fréchet Inception Distance (FID) \cite{heusel2017gans} and Kernel Inception Distance (KID) \cite{binkowski2018demystifying}. Moreover, we evaluate semantic consistency using the CLIP score \cite{radford2021learning} by measuring the cosine similarity between the CLIP features of each generated image and its corresponding ground truth. For each object, we render 4 views using cameras with yaw angles of \{0\textdegree{}, 90\textdegree{}, 180\textdegree{}, 270\textdegree{}\} and a pitch angle of 30\textdegree{}. The camera is positioned with a radius of 2.0 and looks at the origin with a FoV of 40\textdegree{}, consistent with TRELLIS \cite{xiang2024structured}. While FID and KID are calculated between the ground truth and generated sets (6,000 images for Toys4K and 4,120 images for GSO), the CLIP score is calculated in a pair-wise manner, and we report the mean value to evaluate semantic consistency.

\noindent\textbf{(b) Geometry quality}
For 3D geometry evaluation, we adopt Point cloud FID (P-FID) \cite{nichol2022point}, Coverage Score (COV), and Minimum Matching Distance (MMD) using Chamfer Distance (CD). Following previous work \cite{lan2024gaussiananything, xiang2024structured, nichol2022point}, we sampled 4096 points from each GT/generated point cloud, which are obtained from the unprojected multi-view depth maps using the farthest point sampling.

\section{More Results}
\label{sec:sup more results}
In this section, we provide additional qualitative examples and comparison results to further demonstrate the performance of our \mname.

\subsection{Baselines using occluded input}

We have stated in the main paper that "occluded images will lead to incomplete or broken structures" in current 3D generative models. Here, we provide examples where pix2gestalt is omitted and the occluded images are directly used as the input. As shown in Fig.~\ref{fig:occluded input}, when baseline models receive images of partially visible objects as input, they often fail to recover complete and intact 3D assets.

\subsection{More single-view to 3D examples}
\begin{figure*}[ht]
    \centering
    \includegraphics[width=\linewidth]{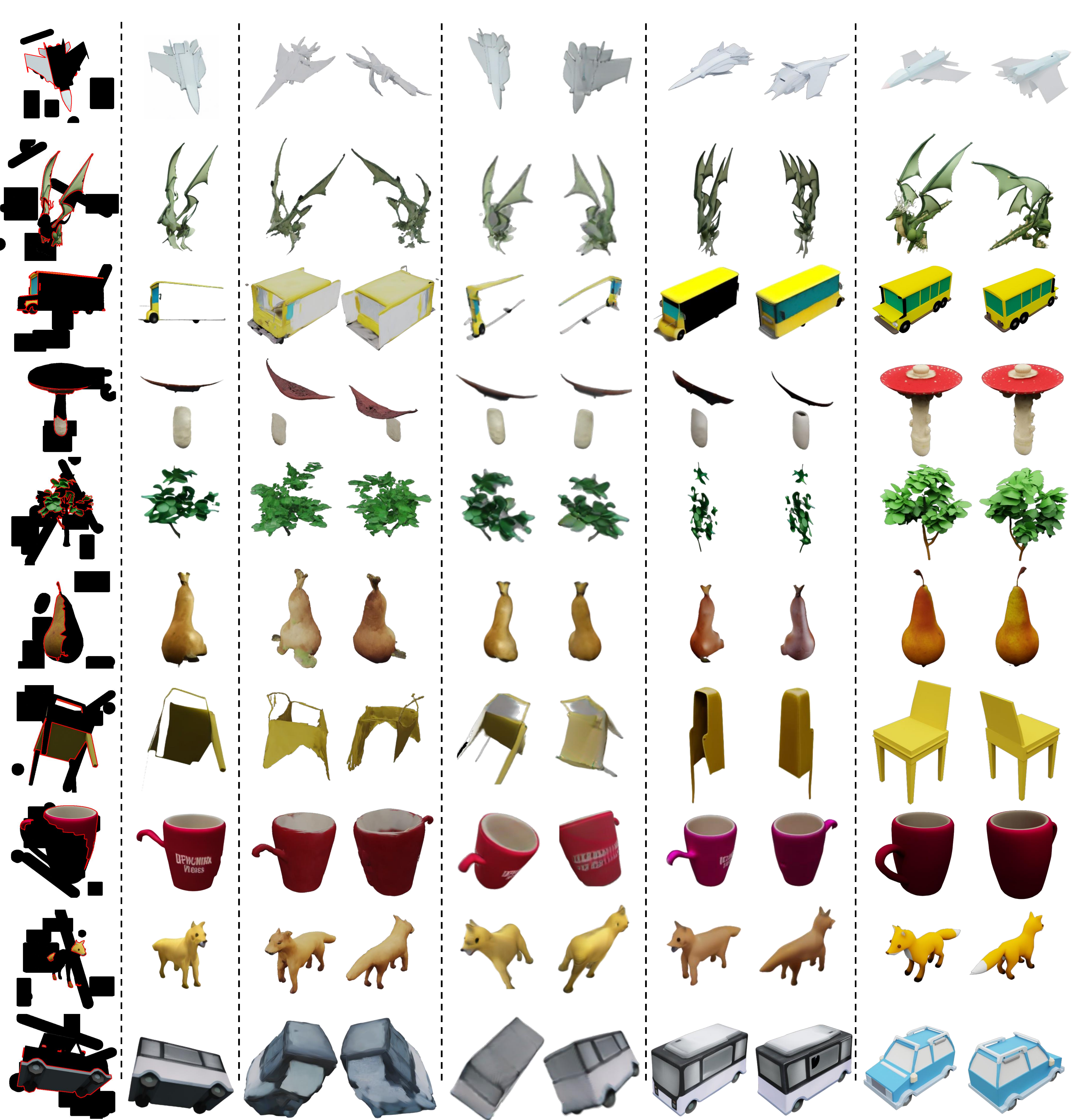}
    \begin{picture}(0,0)
      \put(-250,3){\footnotesize \textbf{Occluded input}}
      \put(-192,3){\footnotesize \textbf{pix2gestalt \cite{ozguroglu2024pix2gestalt}}}
      \put(-130,3){\footnotesize \textbf{GaussianAnything \cite{lan2024gaussiananything}}}
      \put(-20,3){\footnotesize \textbf{Real3D \cite{jiang2024real3d}}}
      \put(75,3){\footnotesize \textbf{TRELLIS \cite{xiang2024structured}}}
      \put(170,3){\footnotesize \textbf{\mname (Ours)}}
    \end{picture}
  
  \caption{\textbf{Additional single-view to 3D comparison examples.} The occluders are shown in black and the visible regions are highlighted with red outlines.}
  \label{fig:sup_single_view}
\end{figure*}

Due to the page restriction, we only provide limited examples in the main paper. Here we visualize more examples of single-view to 3D to further demonstrate the effectiveness of our method in Fig.~\ref{fig:sup_single_view}. The results show that compared with the 2D amodal completion + 3D generation baselines, our \mname yields higher quality 3D assets across multiple categories.

\subsection{More multi-view to 3D examples}

\begin{figure}[ht]
    \centering
    \includegraphics[width=\linewidth]{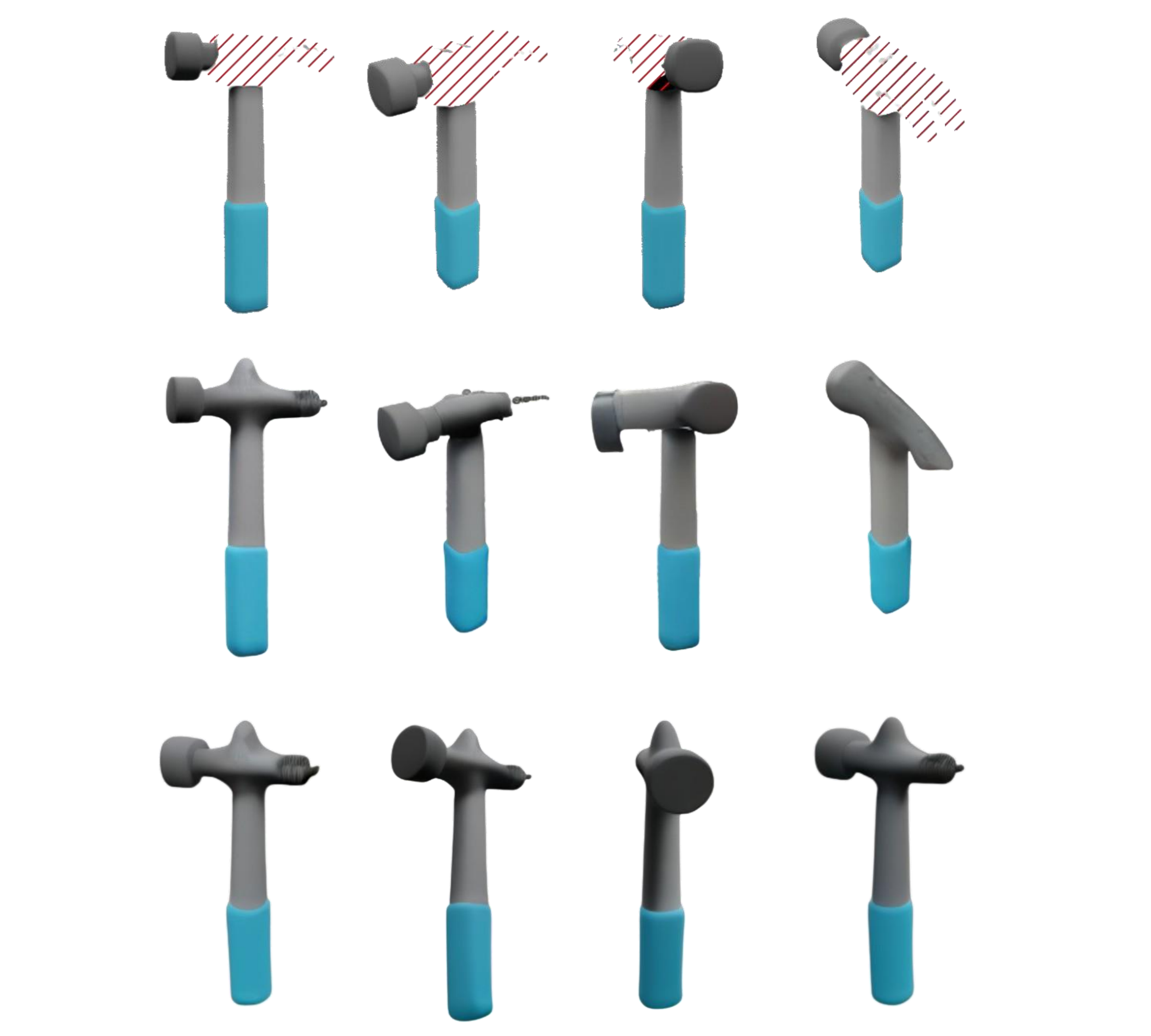}
    \begin{picture}(0,0)
      \put(-50,152){\footnotesize \textbf{Multi-view Occluded input}}
      \put(-25,82){\footnotesize \textbf{pix2gestalt \cite{ozguroglu2024pix2gestalt}}}
      \put(-55,6){\footnotesize \textbf{pix2gestalt \cite{ozguroglu2024pix2gestalt} + Zero123++ \cite{shi2023zero123++}}}
  \end{picture}
    \caption{\textbf{Example of ``pix2gestalt'' and ``pix2gestalt + MV'' input of multi-view to 3D evaluation.}}
    \label{fig:mv setting}
\end{figure}
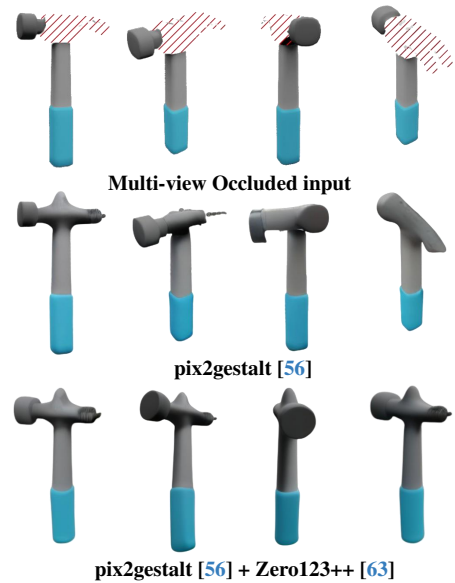

We first provide visualized examples to explicitly explain the difference between the ``pix2gestalt'' and ``pix2gestalt + MV'' settings in the multi-view to 3D generation in Fig.~\ref{fig:mv setting}. For the ``pix2gestalt'' setting, we directly implement pix2gestalt for the amodal completion independently under each view. For the ``pix2gestalt + MV'' setting, we first choose the view with the greatest visibility from the 4 occluded views, then use pix2gestalt to complete the object (which is shown in the pix2gestalt column in the qualitative result), and subsequently use Zero123++ to get the 4 consistent views as the input of LaRa and TRELLIS. It can be observed that pix2gestalt alone results in obvious multi-view inconsistency, while with Zero123++ the consistency is significantly improved, thus leading to better 3D generation quality.

\begin{figure*}[!ht]
  \centering
    \centering
    \includegraphics[width=\linewidth]{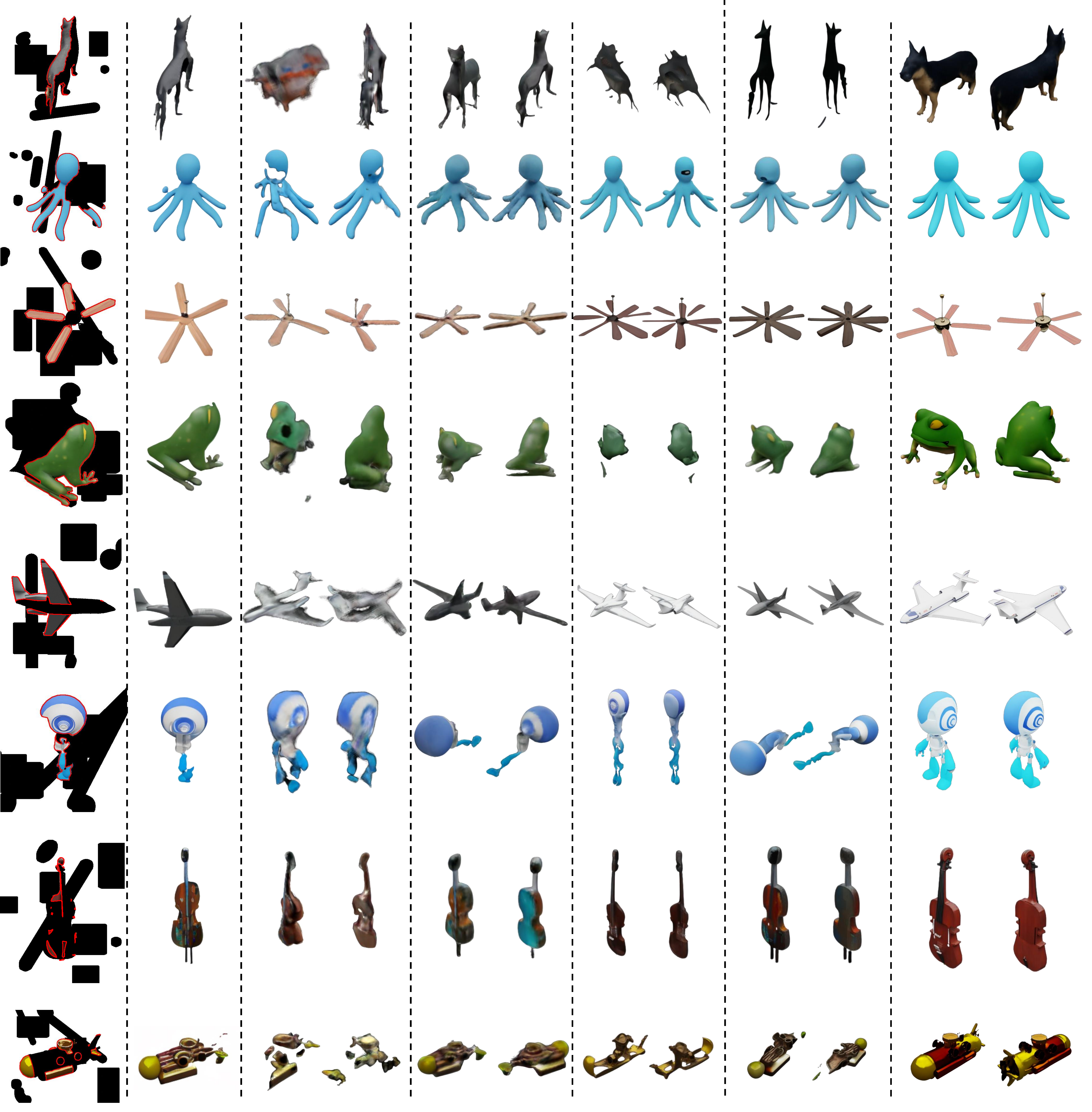}
    \begin{picture}(0,0)
      \put(-250,3){\footnotesize \textbf{Occluded input}}
      \put(-190,3){\footnotesize \textbf{pix2gestalt \cite{ozguroglu2024pix2gestalt}}}
      \put(-113,3){\footnotesize \textbf{LaRa \cite{chen2024lara}}}
      \put(-52,3){\footnotesize \textbf{LaRa \cite{chen2024lara} (+MV)}}
      \put(22,3){\footnotesize \textbf{TRELLIS \cite{xiang2024structured}}}
      \put(85,3){\footnotesize \textbf{TRELLIS \cite{xiang2024structured} (+MV)}}
      \put(175,3){\footnotesize \textbf{\mname (Ours)}}
    \end{picture}
  
  \caption{\textbf{Additional multi-view to 3D comparison examples.} The occluders are shown in black and the visible regions are highlighted with red outlines.}
  \label{fig:sup_multi_view}
\end{figure*}
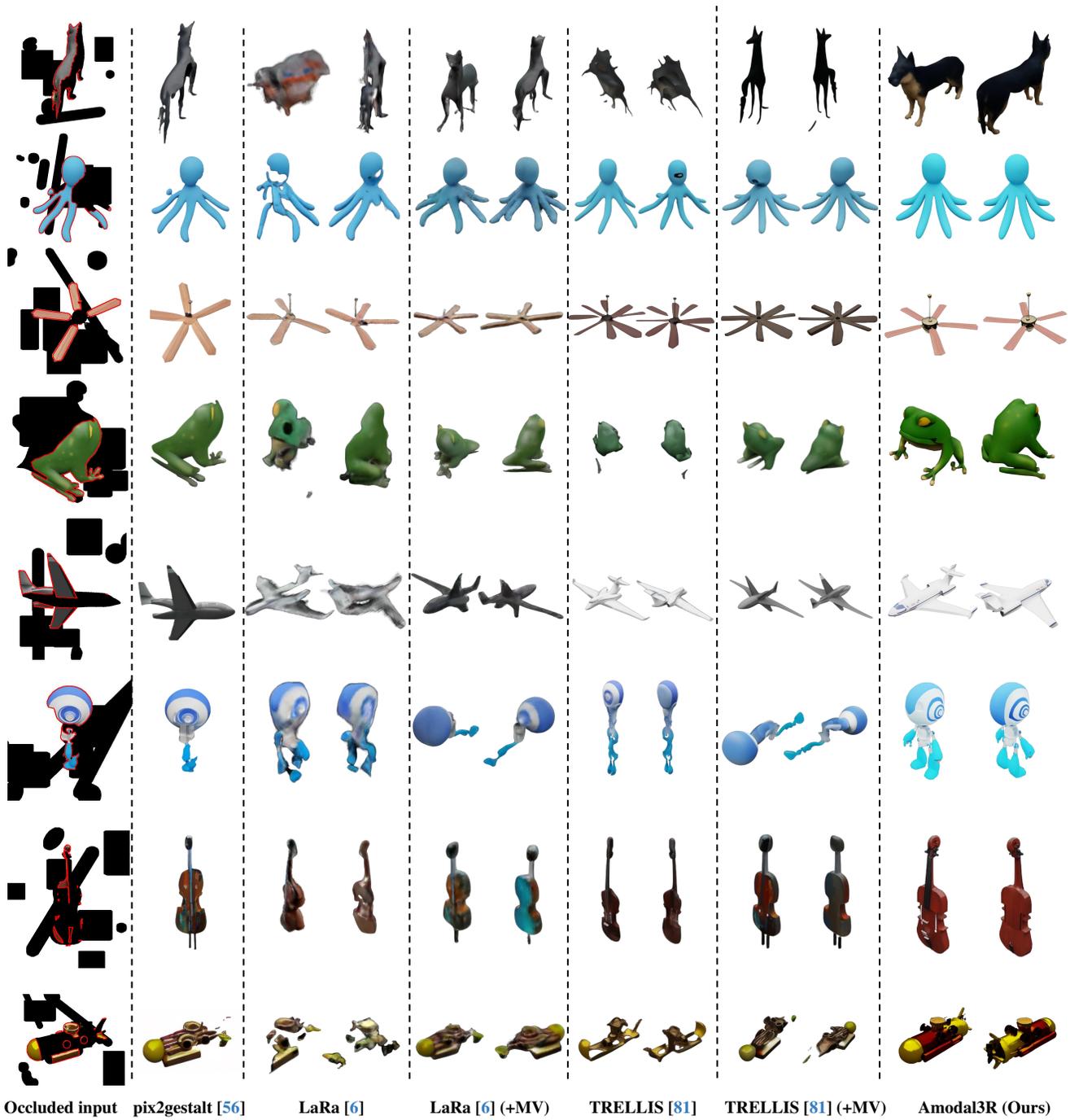

More multi-view to 3D examples are provided in Fig.~\ref{fig:sup_multi_view}, where our \mname again generates 3D assets with better quality than the baselines.

\subsection{More diverse 3D reconstruction results}
\begin{figure*}[!ht]
  \centering
    \centering
    \includegraphics[width=\linewidth]{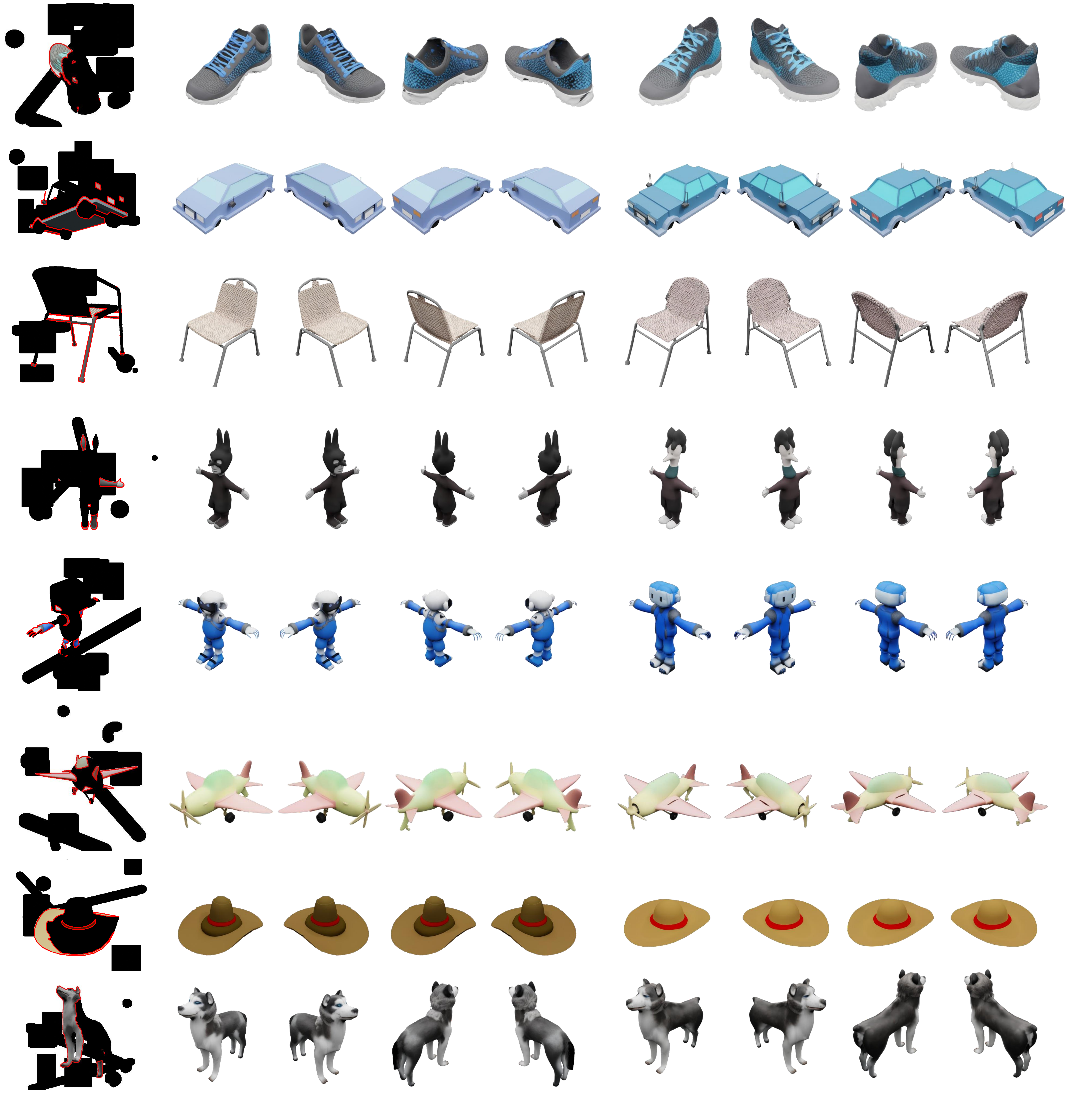}
    \begin{picture}(0,0)
      \put(-238,7){\footnotesize \textbf{Occluded input}}
      \put(-100,7){\footnotesize \textbf{Sample result 1}}
      \put(115,7){\footnotesize \textbf{Sample result 2}}
    \end{picture}
  
  \caption{\textbf{Additional diverse examples.} The occluders are shown in black and the visible regions are highlighted with red outlines.}
  \label{fig:sup_diverse}
\end{figure*}

In Fig.~\ref{fig:sup_diverse}, we show more examples to demonstrate that \mname is able to generate diverse reasonable results from the occluded input with multiple stochastic samplings.

\subsection{More in-the-wild results}
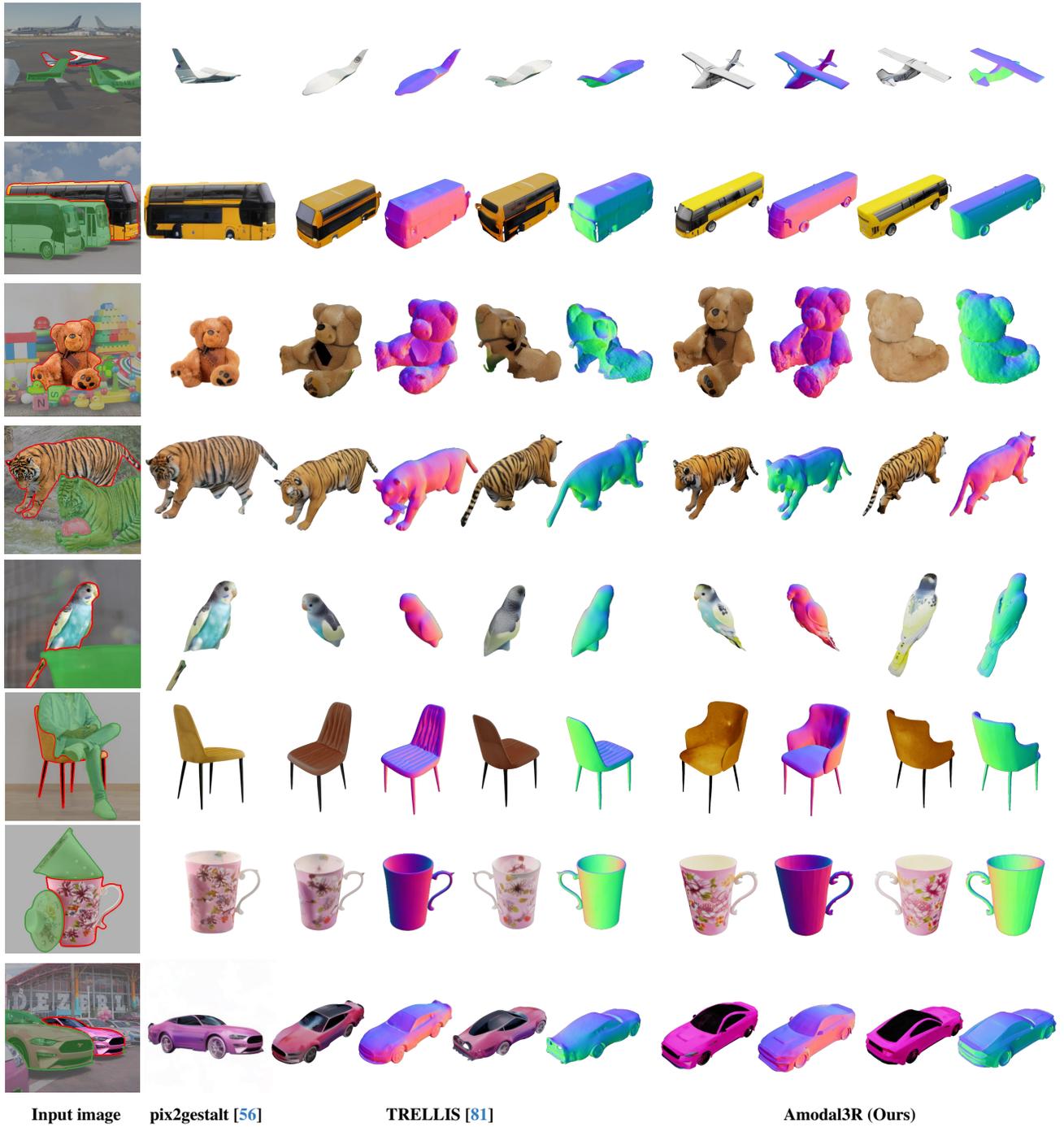
\begin{figure*}[!ht]
    \centering
    \includegraphics[width=\linewidth]{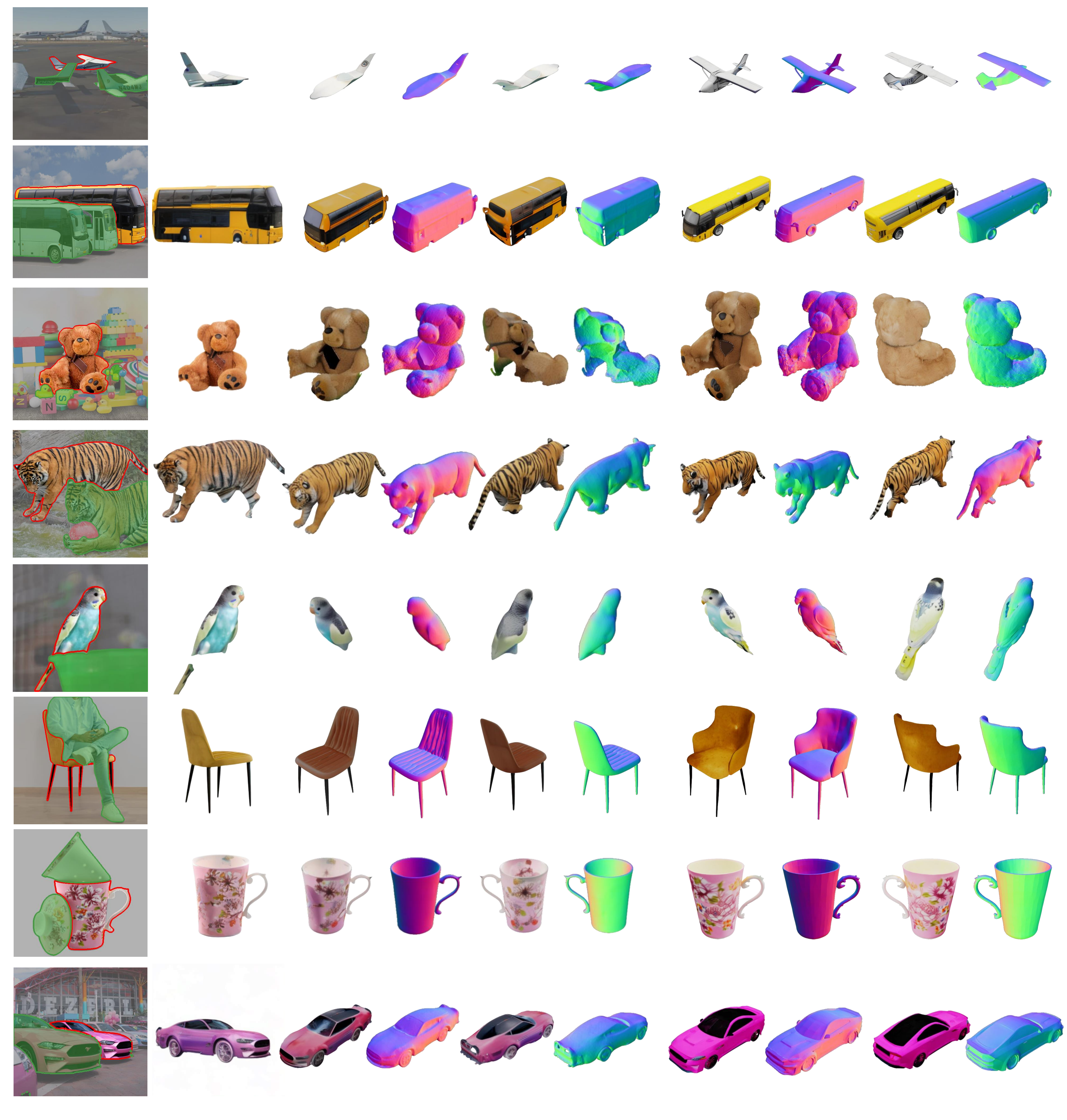}
    \begin{picture}(0,0)
      \put(-230,10){\footnotesize \textbf{Input image}}
      \put(-175,10){\footnotesize \textbf{pix2gestalt \cite{ozguroglu2024pix2gestalt}}}
      \put(-65,10){\footnotesize \textbf{TRELLIS \cite{xiang2024structured}}}
      \put(120,10){\footnotesize \textbf{\mname (Ours)}}
    \end{picture}
  
  \caption{\textbf{Additional in-the-wild examples compared with pix2gestalt + TRELLIS.} The target objects and occluders are marked with the \textcolor[RGB]{250,0,0}{red} and \textcolor[RGB]{68,153,69}{green} outlines.}
  \label{fig:sup_wild_p2g}
\end{figure*}

We provide more examples where we compare \mname with ``pix2gestalt + TRELLIS'' pipeline in \cref{fig:sup_wild_p2g}. The results further demonstrate that 2D amodal method lacks 3D geometric understanding, often resulting in improper completion, such as completing the armrest of the chair as the backrest. In contrast, \mname generates more plausible results with reasonable geometry and rich textural details.

\begin{figure*}[!ht]
    \centering
    \includegraphics[width=\linewidth]{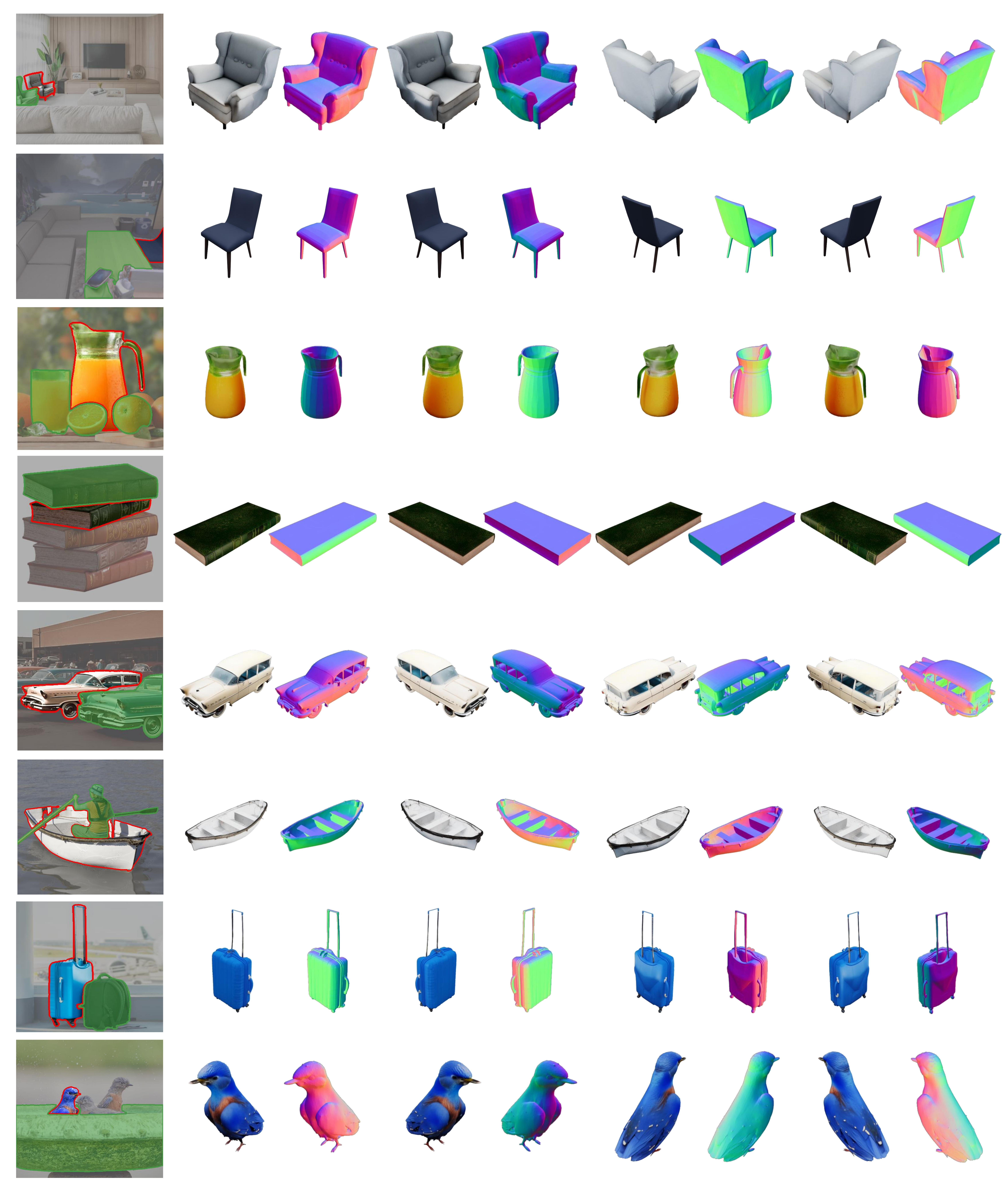}
    \begin{picture}(0,0)
      \put(-225,10){\footnotesize \textbf{Input image}}
      \put(10,10){\footnotesize \textbf{Reconstruction result}}
    \end{picture}
  
  \caption{\textbf{Additional in-the-wild examples.} The target objects and occluders are marked with the \textcolor[RGB]{250,0,0}{red} and \textcolor[RGB]{68,153,69}{green} outlines.}
  \label{fig:sup_wild}
\end{figure*}

\begin{figure*}[!ht]
    \centering
    \includegraphics[width=\linewidth]{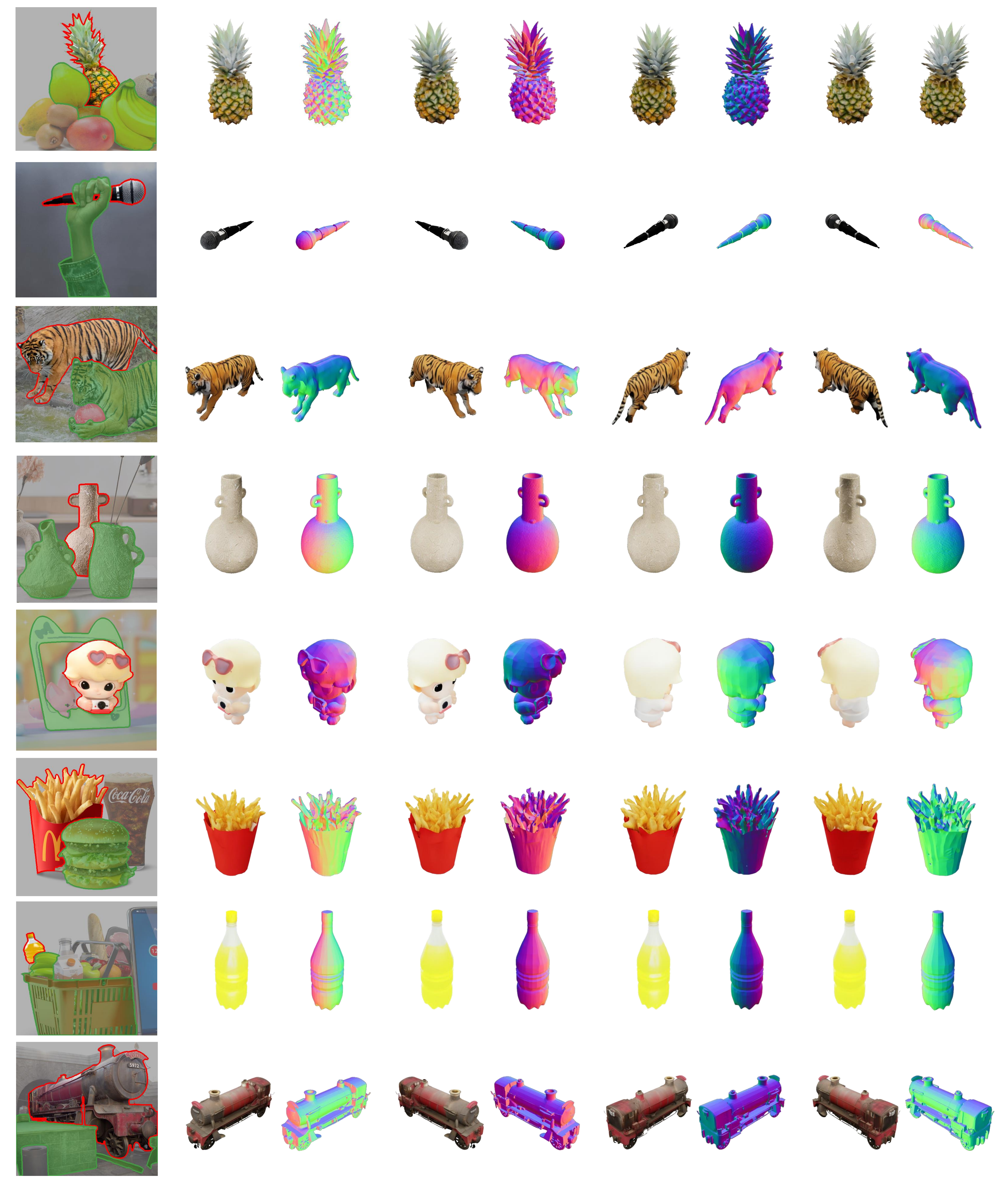}
    \begin{picture}(0,0)
      \put(-225,10){\footnotesize \textbf{Input image}}
      \put(10,10){\footnotesize \textbf{Reconstruction result}}
    \end{picture}
  
  \caption{\textbf{Additional in-the-wild examples.} The target objects and occluders are marked with the \textcolor[RGB]{250,0,0}{red} and \textcolor[RGB]{68,153,69}{green} outlines.}
  \label{fig:sup_wild_2}
\end{figure*}

In Fig.~\ref{fig:sup_wild} and Fig.~\ref{fig:sup_wild_2} we provide more visualization results on the in-the-wild images and the corresponding colored normal maps to show the geometry details. We implement Segment Anything \cite{kirillov2023segment} to obtain the segmentation masks.

\subsection{Discussion, Limitation and Future Work}
While \mname achieves impressive 3D amodal completion, it comes with several limitations we hope to solve in the future. \textbf{1) Dataset expansion.} Due to the computational resources limitation, \mname is trained on very limited data, consisting of only 20,627 synthetic 3D assets, predominantly confined to the furniture category. Consequently, completions on some complex or out-of-distribution objects may fail or lead to unrealistic structures. We believe that training on larger datasets, e.g. Objaverse-XL \cite{deitke2024objaverse}, could mitigate these issues. \textbf{2) Real-World data adaptation}. Different from pix2gestalt, \mname is trained exclusively on synthetic data. As a result, it cannot leverage environmental cues and must rely solely on the visible portions of occluded objects for reconstruction. Creating real-world 3D modal datasets will further enhance the ability to apply models to real scenes. \textbf{3) Controllable completion}. Currently, how objects are completed is entirely up to the model itself and lacks control. Therefore, to further enhance the model to accept additional conditions, such as text, and allow users to edit and control the style of the completion process will be an important research direction for us in the future.

\end{document}